\documentclass[twoside,11pt]{article}
\usepackage{blindtext}

\usepackage[preprint]{jmlr2e}

\usepackage[utf8]{inputenc}
\usepackage[T1]{fontenc}
\usepackage{lmodern}
\usepackage{amsmath}
\usepackage{paralist}
\usepackage{float}
\usepackage{accents}
\usepackage{color}
\usepackage{bm}
\usepackage{lscape}
\usepackage{subfig}
\usepackage{longtable}
\usepackage{multibib}
\usepackage{booktabs}
\newcites{suppl}{References} 
\usepackage{tikz}
\usetikzlibrary{positioning}
\usepackage{dsfont}
\usepackage{pgfplotstable}

\newcommand{\eps}{{\varepsilon}}
\renewcommand{\phi}{\varphi}
\newcommand{\R}{\mathbb{R}}
\newcommand{\Z}{\mathbb{Z}}
\newcommand{\N}{\mathbb{N}}
\newcommand{\pr}{\mathbb{P}}        
\newcommand{\ex}{\mathbb{E}}        
\newcommand{\var}{\textnormal{Var}} 
\newcommand{\cov}{\textnormal{Cov}} 
\newcommand{\Nc}{\mathcal{N}}
\newcommand{\Fc}{\mathcal{F}}
\newcommand{\Ec}{\mathcal{E}}
\newcommand{\Oc}{\mathcal{O}}
\newcommand{\LogReg}{$\mathsf{LogReg}$} 
\newcommand{\NN}{$\mathsf{NN}$}
\newcommand{\NNI}{$\mathsf{NN-init}$}
\newcommand{\NNC}{$\mathsf{NN-cont}$}
\newcommand{\argmin}{\textnormal{argmin}}
\newcommand{\diff}{{\,\mathrm{d}}}
\newcommand{\Gum}{\textnormal{Gum}}           
\newcommand{\iid}{\textnormal{IID}}   
\newcommand{\ma}{\textnormal{MA}}   
\newcommand{\ar}{\textnormal{AR}}   
\newcommand{\convw}{\rightsquigarrow}                           

\newtheorem{assumption}[theorem]{Assumption}

\newcolumntype{H}{>{\setbox0=\hbox\bgroup}c<{\egroup}@{}}
\newcolumntype{Z}{>{\setbox0=\hbox\bgroup}c<{\egroup}@{\hspace*{-\tabcolsep}}}
\allowdisplaybreaks[4]

\definecolor{light-gray}{gray}{0.99}
\pgfplotstableset{
	color cells/.style={
		col sep=comma,
		string type,
		postproc cell content/.code={%
			\pgfkeysalso{@cell 
				content=\rule{0cm}{2.4ex}\cellcolor{black!##1!light-gray}\pgfmathtruncatemacro\number{##1}\ifnum\number>50\color{white}\fi##1}%
		},
		columns/x/.style={
			column name={},
			postproc cell content/.code={}
		}
	}
}

\usepackage{lastpage}
\jmlrheading{24}{2023}{1-\pageref{LastPage}}{9/23}{-}{}{Florian Heinrichs}

\ShortHeadings{Online Monitoring of ML Models}{Heinrichs}
\firstpageno{1}

\begin{document}
\title{Monitoring Machine Learning Models: Online Detection of Relevant Deviations}

\author{\name Florian Heinrichs \email mail@florian-heinrichs.de \\
	\addr SNAP GmbH\\
	Gesundheitscampus-Süd 17\\
	44801 Bochum, Germany}

\editor{-}

\maketitle

\begin{abstract}%
	Machine learning models are essential tools in various domains, but their performance can degrade over time due to changes in data distribution or other factors. On one hand, detecting and addressing such degradations is crucial for maintaining the models' reliability. On the other hand, given enough data, any arbitrary small change of quality can be detected. As interventions, such as model re-training or replacement, can be expensive, we argue that they should only be carried out when changes exceed a given threshold. We propose a sequential monitoring scheme to detect these relevant changes. The proposed method reduces unnecessary alerts and overcomes the multiple testing problem by accounting for temporal dependence of the measured model quality. Conditions for consistency and specified asymptotic levels are provided. Empirical validation using simulated and real data demonstrates the superiority of our approach in detecting relevant changes in model quality compared to benchmark methods. Our research contributes a practical solution for distinguishing between minor fluctuations and meaningful degradations in machine learning model performance, ensuring their reliability in dynamic environments.
\end{abstract}


\maketitle

\section{Introduction} \label{sec:intro}

In recent decades, the field of machine learning has been rapidly evolving. This progress is specifically due to advancements in deep learning, which helped to solve important challenges in computer vision, speech recognition and natural language processing. In particular the development of convolutional neural networks (CNNs) \citep{lecun1989} and long short-term memory (LSTM) networks \citep{hochreiter1997} led to models such as AlexNet and ResNet that won the ImageNet competition and set new benchmarks for image classifiers \citep{krizhevsky2017, he2016}. Further, the attention mechanism and the transformer architecture lead to Google's BERT \citep{devlin2018} and LaMDA \citep{adiwardana2020} and OpenAI's GPT models \citep{radford2018, radford2019, brown2020} that showed impressive performance on tasks related natural language processing. These recent developments in deep learning paved the way for manifold applications of ML-based software in industry. 

However, when ML-based software transitions from academic research and proofs of concept to real-world systems, new complexities are introduced and challenges must be overcome. \cite{sculley2015} first investigated the general technical debts of ML systems. More recently, \cite{vela2022} studied the long-term performance of initially trained neural networks and found that 91\% of the studied models degrade over time. If a model's quality declines unnoticed, its predictions might lead to silent failures with disastrous consequences. To ensure that ML models perform reliably, they need to be monitored and maintained continuously. When a failure is detected, the model might need to be re-trained, updated or completely replaced by another one. Nevertheless, these interventions can be costly, especially if the model is integrated into a highly complex system. Thus, an ideal monitoring scheme would detect a model degradation as soon as it occurs without raising unnecessary alarms. In practice however, this is often not possible and we must keep a balance between false positives (detecting a model degradation when it is indeed stable) and false negatives (overseeing model failures).

Mathematically, let $\mu_1, \dots, \mu_T$ be the model's quality over its life span. Here ``quality'' might refer to the model's accuracy, the confidence of its predictions, or any other metric that quantifies its performance. We could try to detect deviations of $\mu_i$ from the initial value $\mu_1$, i.\,e., raise an alarm at time $i$ if $\mu_i \neq \mu_1$. 

However, given sufficient data, we can detect any arbitrary small change of the model's quality, even if we account for random fluctuations of the observed quality. For example, if a model has an (estimated) initial accuracy of 95\%, we might detect a deviation if the accuracy decreases to 94.9\%, 94.99\%, 94.999\% or any other arbitrary quantity slightly smaller than 95\%. Yet often it is preferable to keep a deployed model as it is instead of putting resources into its re-training or replacement, as long as the decrease of its quality is sufficiently small. Thus, rather than detecting any deviation from the initial value $\mu_1$, we might only be interested to detect \textit{relevant} deviations. More specifically, given a tolerable deviation $\Delta \ge 0$ and a baseline $\mu_0$, we would intervene only if the model's quality deviates from the baseline more than is tolerated, i.\,e., if $|\mu_i -\mu_0| > \Delta$. Using this notion of relevant deviations helps us to avoid rushed reactions while still being able to detect failures of the model. Note that with this definition of relevant deviation, we also detect increases in the model's performance. This might be counterintuitive initially, but if a model's quality improves over time without a plausible reason, we might as well want to get notified and investigate this effect.

In practice, the real ``out-of-sample'' quality $\mu_i$, defined as expected quality with respect to the underlying data distribution, is unknown and we need to estimate it. To account for random fluctuations, it is sensible to use statistical hypothesis tests to test for null hypotheses of the form
\begin{equation} \label{eq:null_single}
	H_0^{(i)}: | \mu_i - \mu _0 | \le \Delta \quad vs.\quad H_1^{(i)}: | \mu_i - \mu _0 | > \Delta,
\end{equation}
for $i=1, \dots, T$. If we use an hypothesis test with some level $\alpha$, that denotes the probability of falsely rejecting the null hypothesis, and repeat it for several time points, we run into the problem of \textit{multiple testing} since the probabilities of falsely rejecting a null hypotheses accumulate. A simple solution is to adapt the significance levels of the single tests to control the overall significance level. However, this approach is expected to be more conservative than a truly sequential monitoring scheme that takes temporal dependencies of the observed data into account.

In the present work, we propose a monitoring scheme for the quality of ML models that is capable to detect relevant deviations. We provide theoretical guarantees that allow to bound the probability of a false alarm and prove that the probability of missing a model degradation vanishes asymptotically. Further, we show that the proposed methodology works through a simulation study and real data experiments. 

The remainder of this paper is structured as follows. Section \ref{sec:rel_work} provides an overview of related work. In Section \ref{sec:main}, we give rigorous definitions of the necessary concepts and present the monitoring scheme and its corresponding theoretical guarantees. In Section \ref{sec:emp_res}, we briefly describe alternative monitoring schemes and benchmark the proposed methods against the alternatives in terms of simulated and real data experiments. Finally, Section \ref{sec:con} concludes the paper.

\section{Related Work} \label{sec:rel_work}

Although the concept of ML model monitoring is not new, it has come into focus of recent research. Since the initial work by \cite{sculley2015}, MLOps, at the intersection of machine learning and DevOps, has been adopted by ML practitioners to maintain ML models in production \citep{klaise2020, testi2022}. Recent studies highlighted the importance of monitoring through structured interviews \citep{haakman2021, shergadwala2022}. \cite{schroder2022} review challenges and methods of monitoring ML models. 

One approach to monitor ML systems is to focus on the data fed into the model. For example, \cite{rukat2020} proposed a methodology to measure and improve the data quality. Further approaches include the detection of drifts in the data. \cite{storkey2009} characterizes different sources of data drift. A particular attention is classically given to concept drift, see \cite{gama2014, lu2018} for recent reviews.

Another approach is to monitor the model directly. If there is a feedback loop and we get the ground truth for a prediction at some point in the future, we can use this information to calculate the model's accuracy or a similar metric. Contrarily, if we do not get any feedback, we can monitor confidences of the predictions to detect changes in the model's quality. In a recent work, \cite{vela2022} studied the degradation of ML models and compared it with related concepts, such as concept drift.

Monitoring the quality of a machine learning model can be generalized to monitoring any real-valued quantity over time. Thus it is sensible to take into account methodology from the field of statistics that deals with this type of data, namely time series analysis and, more specifically, methods related to change point detection. 

For \textit{classic hypotheses}, i.\,e., for $\Delta=0$ in \eqref{eq:null_single}, several online monitoring schemes based on the CUSUM statistic have been proposed \citep{lai1995, chu1996, kirch2022, gosmann2022}. Further, \cite{chen2019} studied a monitoring scheme based on nearest neighbors. Finally, monitoring schemes for time series that fit specific models, such as the GARCH model, have been proposed \citep{berkes2004}.

When it comes to relevant hypotheses with $\Delta\ge 0$, to the best of our knowledge, no online monitoring schemes have been proposed. Yet, different approaches of testing for relevant changes retrospectively have been studied. \cite{dette2019} estimate the time during which the relevant threshold $\Delta$ is exceeded and reject the null hypothesis of no change when the duration is sufficiently long. Using self-normalization, \cite{dette2020} proposed a test statistic for functional data, whereas \cite{heinrichs2021} tested for deviations in the $L^2$-norm. Finally, \cite{bucher2021} proposed a retrospective test for the hypotheses in \eqref{eq:null_single} based on an estimator of the function $\mu$.

In contrast to existing work, we propose an online monitoring scheme for relevant hypotheses and use this scheme to monitor the quality of ML models. Note that, although we refer to $\mu$ as the model quality, it could be any quantity related to the reliability of the entire ML-based system. In particular, it could be some ``quality metric'' to monitor the quality of the data.

\section{Methodology} 
\label{sec:main}

In this section, we formalize the testing problem, introduce two monitoring schemes and provide theoretical guarantees for their behavior.

\subsection{The General Testing Problem} 

In the following, we assume that the quality of an ML model is observed over its life span of $T$ time steps with a frequency of $n$ observations per time step, yielding the model
$$ X_{i, n} = \mu(i/n) + \eps_i, \quad i=1,\dots, T\cdot n, $$
where $\mu:[0, T]\to\R$ denotes the unknown quality function and $\{\eps_i\}_{i\in\N}$ is a stationary sequence of centered random variables. For example, if the quality is measured daily over a period of 12 months, $n=30$ and $T=12$. Note that we do not assume independence of the error terms and allow for temporal dependencies that are common in many applications.

We are interested in detecting changes in the quality function $\mu$ over time. The classic approach would be to test for the null hypothesis
\begin{equation*}
H_0: \mu(t) = \mu(0),
\end{equation*}
at time point $t\in[0,T]$. However, the quality of an ML model changes eventually, and given sufficient data, we are able to detect any change of the quality function $\mu$. To focus on deviations of practical significance, we introduce the concept of a maximal tolerable deviation, denoted as $\Delta\ge 0$, and formulate tests for the relevant hypotheses
\begin{equation*}
H_0^{(t)}: |\mu(t) - \mu(0)| \le \Delta \quad vs. \quad H_1^{(t)}: |\mu(t) - \mu(0)| > \Delta,
\end{equation*}
for $t\in [0, T]$, where the classic null hypothesis is covered with $\Delta = 0$. In many cases, our interest extends beyond the initial value $\mu(0)$ to encompass a ``target'' quantity denoted as $g(\mu)$. For example, we might want to have an ML model with an accuracy of 95\% and be able to detect any deviation larger than 1\%, so $g(\mu) = 95\%$ and $\Delta = 1\%$. In another application we might know that the accuracy is (approximately) constant during the first period, but do not care too much about its actual value. In this case, $g(\mu) = \int_0^1 \mu(x)\diff x$. Thus, given a target $g(\mu)$ and a threshold for the maximal tolerable deviations $\Delta \ge 0$, we are interested in hypotheses of the form
\begin{equation}\label{eq:null}
H_0^{(t)}: |\mu(t) - g(\mu)| \le \Delta \quad vs. \quad H_1^{(t)}: |\mu(t) - g(\mu)| > \Delta,
\end{equation}
for $t\in[0,T]$. As indicated by the notation, $g$ might be an arbitrary functional of $\mu$.

When monitoring the quality of an ML model, our goal is not limited to testing a single hypothesis $H_0^{(t)}$ at a specific time point $t$, but rather across any time point within the interval $[0, T]$. As pointed out earlier, employing a hypothesis test with a significance level of $\alpha$ and repeating it for multiple time points gives rise to the challenge of \textit{multiple testing}, whereby the probabilities of falsely rejecting null hypotheses accumulate. In order to tackle this issue, a straightforward approach consists of modifying the significance levels of individual tests to manage the overall significance level. Nevertheless, it is crucial to acknowledge that this method is anticipated to be more conservative than a genuinely sequential monitoring scheme that considers the temporal dependencies inherent in the observed data. 

To find a balance between false alarms and failing to detect a decline in the model quality, we extend the concepts of significance levels and consistency to sequential monitoring schemes. More specifically, we refer to a sequence of decision rules for the hypotheses in \eqref{eq:null} as \textit{monitoring scheme with asymptotic level $\alpha$}, if the probability of falsely rejecting any null hypothesis is asymptotically bounded by $\alpha \in (0, 1)$ under the joint null hypothesis $\overline{H}_0^{(T)} = \bigcap_{t\in[0, T]} H_0^{(t)}$, i.\,e., if
$$ \limsup_{n\to\infty}\pr(\text{Reject}~\overline{H}_0^{(T)} | \overline{H}_0^{(T)}) \le \alpha. $$

Further the monitoring scheme is \textit{consistent} if 
$$ \lim_{n\to\infty}\pr(\text{Reject}~\overline{H}_0^{(T)} | \overline{H}_1^{(T)}) = 1. $$

\subsection{A New Monitoring Scheme Based on the Maximal Deviation} \label{sec:mon_scheme}

Subsequently, we assume that the unknown quality function $\mu$ is sufficiently smooth and the errors $\{\eps_i\}_{i\in\N}$ exhibit a form of ``weak'' dependence, as elaborated in Section \ref{sec:theory}. These two properties allow us to estimate $\mu$ based on the observations $X_{1,n}, \dots, X_{Tn, n}$, which can be done through local linear regression. More specifically, let $h_n$ be a positive bandwidth that vanishes as $n\to\infty$. Further, given a kernel function $K$, define $K_h(\cdot) = K(\frac{\cdot}{h})$.  Then, the \textit{local linear estimator} $\hat{\mu}_{h_n}$ with bandwidth $h_n$ is defined as the first coordinate of
$$ \big(\hat{\mu}_{h_n}, \widehat{\mu'}_{h_n}(t)\big) = \argmin_{b_0, b_1\in\R} \sum_{i=1}^{Tn} \bigg(X_{i, n} - b_0 - b_1 \Big( \frac{i}{n}-t\Big)\bigg)^2 K_{h_n}\Big(\frac{i}{n}-t\Big), $$ 
see, for example, \cite{fan1996}. Since $\hat{\mu}_{h_n}$ is a biased estimator of $\mu$, we use the Jackknife estimator $\tilde{\mu}_{h_n}(t) = 2 \hat{\mu}_{h_n/\sqrt{2}}(t) - \hat{\mu}_{h_n}(t)$, as proposed by \cite{schucany1977}, to reduce this bias.

If the baseline $g(\mu)$ is an unknown quantity, such as the initial quality $\mu(0)$, rather than a fixed quantity, it has to be approximated as well. A valid strategy is to plug $\tilde{\mu}_{h_n}$ into the functional $g$ and obtain the estimator $g(\tilde{\mu}_{h_n})$, but often simpler estimators exist. For example, if $g(\mu)=\mu(0)$, valid estimators are the plug-in estimator $\hat{\mu}_{h_n}(0)$ or simply $\bar{X}_{m_n}=\frac{1}{m_n}\sum_{i=1}^{m_n} X_{i,n}$ for some sequence $m_n=o(n)$ that tends to infinity. The baseline $g(\mu) = \int_0^1 \mu(t)\diff t$ could be estimated by the plug-in estimator $\int_0^1 \hat{\mu}_{h_n}(t) \diff t$ or directly by $\bar{X}_n = \frac{1}{n}\sum_{i=1}^{n} X_{i,n}$. In the following, we only assume that the estimator $\hat{g}_n$ converges (sufficiently fast) to $g(\mu)$. 

To construct a reliable monitoring scheme, we need to take the serial dependence of the errors $\eps_1, \dots, \eps_{Tn}$ into account. The \textit{long-run variance} measures the dependence of these errors and plays a role similar to the variance in case of independent data. For a stationary sequence of random variables $\{\eps_t\}_{t\in\Z}$, the long-run variance is defined as
\begin{equation*}
	\sigma_{lrv}^2 = \sum_{t=-\infty}^{\infty}\cov(\eps_t, \eps_0),
\end{equation*}
where $\sigma_{lrv}^2 = \var(\eps_0)$ if the random variables are uncorrelated. As the long-run variance of the error process is unknown, but contains important information for the monitoring scheme, we need to estimate it. Therefore, let $m_n$ be a sequence proportional to $n^{1/3}$ and $S_{j,k}= \sum_{i=j}^k X_{i,n}$. Then,  
$$\hat{\sigma}_{lrv}^2 = \frac{1}{\lfloor n/m_n\rfloor - 1}\sum_{i=1}^{\lfloor n/m_n\rfloor - 1} \frac{(S_{(j-1)m_n+1,jm_n} - S_{jm_n+1,(j+1)m_n})^2}{2m_n}$$ 
is a consistent estimator of $\sigma_{lrv}^2$, see \cite{wu2007}. 

Further, the monitoring scheme needs to consider the asymptotic behavior of $\tilde{\mu}_{h_n}(t)$. Let $K^*(x) = 2\sqrt{2}K(\sqrt{2}x)-K(x)$ be the ``Jackknife'' version of the kernel $K$, $(K^*)'$ its derivative and $\|K^*\|_2$ denote its $L^2$-norm, i.\,e., $\|K^*\|_2^2 = \int_0^1  \big(K^*\big)^2(x)\diff x$. Then, the asymptotic behavior of $\tilde{\mu}_{h_n}(t)$ is determined by the scaling sequence
$$ \ell_n = \sqrt{2 \log\Big(\frac{T \|(K^*)'\|_2}{2 \pi h_n \|K^*\|_2}\Big)}. $$

Finally, let $\Gum_a$ denote the Gumbel distribution with location parameter $a\in\R$ with cumulative distribution function $\Gum_a((-\infty, x])=\exp\{-\exp\big(-(x-a)\big)\}$ and $q_{a,1-\alpha}$ its $(1-\alpha)$ quantile. Then, we propose to reject the null hypothesis $H_0^{(t)}$, for $t\in[0, T]$, whenever
\begin{equation}\label{eq:mon_scheme}
\big|\tilde{\mu}_{h_n}(t)-\hat{g}_n\big| > \Delta + \big(q_{a,1-\alpha}+\ell_n^2\big) \frac{\hat{\sigma}_{lrv}\|K^*\|_2}{\sqrt{nh_n}\ell_n},
\end{equation}
where $a = \log(2)$ if $\Delta = 0$ and $a = 0$ otherwise.

\begin{corollary}
	Under the assumptions of Theorem \ref{thm:main}, the monitoring scheme defined in \eqref{eq:mon_scheme} is consistent and has asymptotically level $\alpha$.
\end{corollary}

\subsection{Theoretical Guarantees} \label{sec:theory}

As previously mentioned, we allow dependencies between the errors $\{\eps_t\}_{t\in\Z}$. However, in time series analysis assumptions about the dependence structure are necessary to develop valid statistical methods and make meaningful inferences. Different approaches exist to control dependencies, and in this context, we will use the concept of \textit{physical dependence} \citep{Wu2005}, which offers a more intuitive understanding compared to other measures of dependence.

Therefore, let $\eta = \{\eta_i\}_{i\in\Z}$ be a sequence of independent identically distributed random variables and let $\Fc_i$ denote the sequence $(\dots,\eta_{-1}, \eta_0, \eta_1, \dots, \eta_i)$. Further, let $\Fc_i^*$ denote the sequence $(\dots,\eta_{-1}, \eta_0', \eta_1, \dots, \eta_i)$, where $\eta_0$ was replaced by an independent copy $\eta_0'$. Finally, let $G:\R^\N \to \R$ be a possibly nonlinear function such that $G(\Fc_i)$ and $G(\Fc_i^*)$ are properly defined random variables. Then, the physical dependence measure of $G$ with $\ex |G(\Fc_i)|^q < \infty$, for some $q\in\N$, is defined as 
$$ \delta_q(G, i) = \big( \ex | G(\Fc_i) - G(\Fc_i^*) |^q\big)^{1/q},\quad i\in\N. $$

The term $\delta_q(G, i)$ quantifies the influence of $\eta_0$ on $G(\Fc_i)$ and can be interpreted as a measure of the serial dependence of $\{G(\Fc_t)\}_t$ at lag $i$. For instance, if $G$ maps the sequence $(\dots,\eta_{-1}, \eta_0, \eta_1, \dots, \eta_i)$ to $\eta_i$, $G(\Fc_i) = \eta_i$ is independent of $\eta_0'$ for $i\ge 1$ and $\delta_q(G, i) = 0$. In the following, we allow for serial dependence as long as the quantity $\delta_q(G, i)$ vanishes for $i\to\infty$.

We now specify the necessary conditions, which ensure that the decision rule in \eqref{eq:mon_scheme} has the desired properties.

\begin{assumption}\label{ass:main}
	\begin{enumerate}
		\item The function $K:\R\to\R$ is symmetric, twice differentiable, vanishes outside of the interval $[-1, 1]$ and satisfies $\int_{-1}^1 K(x) \diff x = 1$.
		\item The bandwidth $h_n$ converges to zero as $n\to\infty$ with
		$$ nh_n\to\infty, \quad nh_n^7 |\log(h_n)|\to 0,\quad \limsup_{n\to\infty}\tfrac{|\log(h_n)|\log^4n}{n^{1/2}h_n}<\infty. $$
		\item The quality function $\mu$ is twice differentiable and its second derivative is Lipschitz continuous.
		\item The estimator $\hat{g}_n$ of the baseline $g(\mu)$ satisfies 
		$$|\hat{g}_n - g(\mu)|=o_\pr\Big(\frac{1}{\sqrt{nh_n |\log(h_n)|}}\Big),$$
		as $n\to\infty$.
		\item There exists a function $G:\R^\N\to\R$ and a constant $\gamma \in (0,1)$, such that the centered error process $\{\eps_i\}_{i\in\Z}$ can be represented as $\eps_i = G(\Fc_i)$ and $\delta_4(G, i)=\Oc(\gamma^i)$, as $i\to\infty$. The \textit{long-run variance} of $\{\eps_i\}_{i\in\Z}$ exists and is positive.
	\end{enumerate}
\end{assumption}

The conditions in the previous assumption are relatively mild. (1) and (2) are design choices when estimating the underlying quality function $\mu$. The smoothness of the estimate is rather determined by the bandwidth $h_n$ than by the specific choice of $K$. More specifically, (2) holds true if $h_n = n^{-c}$ for some constant $c\in(\tfrac{1}{7},\tfrac{1}{2})$. Note that the choice of $h_n$ is crucial to avoid overfitting and oversmoothing, thus using cross validation to tune the bandwidth is recommended in practice. (3) is a reasonable assumption if the model quality changes smoothly over time. Clearly it does not hold, if the quality changes abruptly at a certain time point. As shown in Section \ref{sec:emp_res}, the proposed methodology is robust if this assumption is not met. Part (4) of Assumption \ref{ass:main} is satisfied by many common estimators, particularly by those mentioned in Section \ref{sec:mon_scheme}. Finally, (5) is a common assumption to control the degree of dependence of a time series and satisfied by many time series of interest. The physical dependence measure is similar to mixing coefficients, yet often easier to bound. 

\begin{theorem}\label{thm:main}
	Let Assumption \ref{ass:main} be satisfied, $\Delta > 0$ a threshold and $\alpha\in(0,1)$. Further, if $\Ec = \{t\in [0, T] : |\mu(t)| = \Delta \}$ is not empty, let $\inf_{t\in\Ec}|\mu''(t)|> 0$.
	
	 Under the joint null hypothesis $\overline{H}_0^{(T)}$, i.\,e., if $ |\mu(t) - g(\mu) | \le \Delta$ for any $t \in [0, T]$, it holds.
	
	$$ \limsup_{n\to\infty} \pr\bigg(\big|\tilde{\mu}_{h_n}(t)-\hat{g}_n\big| > \Delta + \big(q_{0,1-\alpha}+\ell_n^2\big) \frac{\hat{\sigma}_{lrv}\|K^*\|_2}{\sqrt{nh_n}\ell_n}~\text{for some}~t\in[0, T]\bigg) \leq \alpha.$$
	
	Further,
	$$ \lim_{n\to\infty} \pr\bigg(\big|\tilde{\mu}_{h_n}(t)-\hat{g}_n\big| > \Delta + \big(q_{0,1-\alpha}+\ell_n^2\big) \frac{\hat{\sigma}_{lrv}\|K^*\|_2}{\sqrt{nh_n}\ell_n}~\text{for some}~t\in[0, T]\bigg) = 1,$$	
	under the alternative $\overline{H}_1^{(T)}$, i.\,e., if $ |\mu(t) - g(\mu) | > \Delta$ for some $t \in [0, T]$. For $\Delta=0$, the equalities hold if the quantile $q_{0, 1-\alpha}$ is replaced by $q_{\log(2), 1-\alpha}$.
\end{theorem}

A proof of this theorem is provided in Appendix \ref{sec:app_theory}.

A crucial step in the theorem's proof is a Gaussian approximation of $\big|\tilde{\mu}_{h_n}(t)-\hat{g}_n\big|$, where the supremum of this Gaussian approximation converges to a Gumbel distribution. The latter convergence is rather slow and instead of using quantiles of the limiting Gumbel distribution, it is also possible to use approximated quantiles. More specifically, let $\{V_i\}_{i\in\N}$ be a sequence of independent standard normally distributed random variables. If $\Delta > 0$, we can replace $q_{0, 1-\alpha}$ in \eqref{eq:mon_scheme} by the $(1-\alpha)$ quantile of 
$$ G_n = \ell_n \bigg(\sup_{t\in[0, T]} \frac{1}{\|K^*\|_2 \sqrt{nh_n}} \sum_{i=1}^{Tn} V_i K_{h_n}^*\big(\tfrac{i}{n}-t\big) - \ell_n \bigg) $$
to obtain a consistent monitoring scheme with asymptotic level $\alpha$. Similarly, for $\Delta = 0$, we can replace $q_{\log(2),1-\alpha}$ by the $(1-\alpha)$ quantile of 
$$ G_n' = \ell_n \bigg(\sup_{t\in[0, T]} \frac{1}{\|K^*\|_2 \sqrt{nh_n}} \bigg|\sum_{i=1}^{Tn} V_i K_{h_n}^*\big(\tfrac{i}{n}-t\big)\bigg| - \ell_n \bigg). $$
These quantiles can be easily approximated by simulating sequences of independent standard normally distributed random variables $\{V_i\}_{i\in\N}$. Thus, we can modify the monitoring scheme in \eqref{eq:mon_scheme} and reject the null hypothesis whenever
\begin{equation}\label{eq:approx_mon_scheme}
\big|\tilde{\mu}_{h_n}(t)-\hat{g}_n\big| > \Delta + \big(\hat{q}_{1-\alpha}+\ell_n^2\big) \frac{\hat{\sigma}_{lrv}\|K^*\|_2}{\sqrt{nh_n}\ell_n},
\end{equation}
where $\hat{q}_{1-\alpha}$ denotes the $(1-\alpha)$ quantile of $G_n'$ if $\Delta = 0$ and the corresponding quantile of $G_n$ otherwise.

\section{Empirical Results}  \label{sec:emp_res}

In the following we investigate the finite sample properties of the proposed monitoring schemes and compare it with four alternatives. These monitoring schemes are based on different assumptions. For a fair comparison, we use the baseline $g(\mu)=\int_0^1 \mu(t)\diff t$ and monitor the model quality for $t\in[1, T]$. In Section \ref{sec:sim_study}, we test the different methods with simulated data while we train and monitor several models for stationary and non-stationary data in Section \ref{sec:real_data_exp}.

When applying any monitoring scheme, we are not only interested in correctly accepting or rejecting the joint null hypothesis $\overline{H}_0^{(T)}$, but want to detect a relevant deviation as soon as it occurs. Besides the empirical rejection rates, we also compare the time points of rejecting the joint null hypothesis.

\subsection{Alternative Monitoring Schemes}

We provide a brief overview of the monitoring schemes that are compared to the proposed method.

\subsubsection{A Naive Monitoring Scheme}

Given observed values $X_{1, n}, \dots, X_{Tn, n}$, one could simply reject the null hypothesis at point $k$ whenever
\begin{equation}\label{eq:naive_scheme}
\big|X_{k, n} - \bar{X}_n \big| > \Delta,
\end{equation}

where $\bar{X}_n = \frac{1}{n}\sum_{i=1}^n X_{i, n}$. This naive monitoring scheme is clearly not robust towards random fluctuations of the model quality and does not have level $\alpha$ at the boundary of the null hypothesis, i.\,e., when $\sup_{t\in[1, T]} |\mu(t) - g(\mu) |=\Delta$. It can be expected that the false positive rate remains high even for larger deviations.

\subsubsection{$t$-Test-based Monitoring Schemes}

The traditional approach to test if the mean of a sample equals a specific value is to use Student's $t$-test. As long as some variant of the central limit theorem holds, we can generalize the $t$-test to relevant hypotheses of the form \eqref{eq:null} by rejecting the null hypothesis at time $k$ whenever

\begin{equation}\label{eq:t_test}
\bigg| \frac{1}{n} \sum_{i=k+1}^{k+n} \big(X_{i, n} - \bar{X}_n\big) \bigg| > \Delta + q^{(t)}_{1-\alpha} \frac{\hat{\sigma}_{n,1}}{\sqrt{n}},
\end{equation}

where $\hat{\sigma}_{n,1}^2$ denotes an estimator of the variance of $\{X_{i, n } - \bar{X}_n\}_{i=k+1}^{k+n}$ and $q^{(t)}_{1-\alpha}$ denotes the $(1-\alpha)$ quantile of the standard normal distribution in case of $\Delta > 0$ and the $(1-\alpha/2)$ quantile of the same distribution for $\Delta = 0$. 

This formulation of the test does not take into account the multiple testing problem and might lead to a monitoring scheme that does not have asymptotically level $\alpha$. To account for the accumulation of error probabilities, we can replace the individual level $\alpha$ by $\alpha' = \frac{\alpha}{(T-1)n}$ and reject the null hypothesis whenever 

\begin{equation}\label{eq:t_test_corr}
\bigg| \frac{1}{n} \sum_{i=k+1}^{k+n} \big(X_{i, n} - \bar{X}_n\big) \bigg| > \Delta + q^{(t)}_{1-\alpha'} \frac{\hat{\sigma}_{n,1}}{\sqrt{n}}.
\end{equation}

This latter test does not take any dependence of the observed model quality into account and might be overly conservative.

\subsubsection{CUSUM-based Monitoring Schemes for Classic Hypotheses}

The final monitoring schemes are based on the CUSUM statistic, defined as 
$$\Gamma(n, k) =\frac{k}{n}\sum_{i=1}^{n} X_{i,n} - \sum_{i=n+1}^{n+k} X_{i,n},$$ 
which is a classic statistic in the context of change point analysis and the detection of structural breaks. Let 
$$\hat{\sigma}_{n,2}^2 = \frac{1}{n-1} \sum_{i=1}^n \big(X_{i,n}-\bar{X}_n\big)^2$$
be an estimator of the variance of the residuals $\{X_{i,n}-\bar{X}_n\}_{i=1}^{n}$. Further, let $\{W(t)\}_{t>0}$ denote a standard Brownian motion and $q^{(c)}_{1-\alpha}$ be the $(1-\alpha)$ quantile of $\sup_{0<t<1} |W(t)|$. Then, we can reject the null hypothesis whenever

\begin{equation}\label{eq:cusum_test}
\frac{1}{\hat{\sigma}_{n,2}} \sup_{k\ge 1} \frac{\sqrt{n}}{n+k}|\Gamma(n, k)| > q^{(c)}_{1-\alpha}.
\end{equation}

We can replace the classic CUSUM statistic by 
$\sup_{0\le \ell\le k} |\Gamma(n, k) - \Gamma(n, l)|$, which is often referred to as the \textit{Page-CUSUM} statistic \citep{fremdt2015}. More precisely, if $q^{(p)}_{1-\alpha}$ denotes the $(1-\alpha)$ quantile of $\sup_{0<t<1} \sup_{0<s\le t} \Big|W(t) - \frac{1-t}{1-s}W(s)\Big|$, we can reject the null hypothesis whenever 
\begin{equation}\label{eq:page_cusum_test}
\frac{1}{\hat{\sigma}_{n,2}} \sup_{k\ge 1} \frac{\sqrt{n}}{n+k}\sup_{0\le \ell\le k} |\Gamma(n, k) - \Gamma(n, l)| > q^{(p)}_{1-\alpha}.
\end{equation}

By Corollary 2 of \cite{kirch2022}, the monitoring schemes defined by \eqref{eq:cusum_test} and \eqref{eq:page_cusum_test} are consistent with asymptotic level $\alpha$ if the mean is constant over the first $n$ observations, i.\,e., 
\begin{equation*}
\mu(t)\equiv \mu(0)~\text{for}~t\in[0, 1],
\end{equation*} and under some additional assumptions. More specifically, it holds 
$$ \frac{1}{\hat{\sigma}_{n,2}}\sup_{k\ge 1} \frac{\sqrt{n}}{n+k}|\Gamma(n, k)| \convw \sup_{0<t<1} |W(t)| $$
and 
$$ \frac{1}{\hat{\sigma}_{n,2}}\sup_{k\ge 1} \frac{\sqrt{n}}{n+k} \sup_{0\le \ell\le k} |\Gamma(n, k) - \Gamma(n, l)| \convw \sup_{0<t<1} \sup_{0<s\le t} \Big|W(t) - \frac{1-t}{1-s}W(s)\Big|, $$
where ``$\convw$'' denotes weak convergence.

However, the monitoring schemes can be used only to test for the classic hypothesis of $\Delta=0$.

\subsection{Simulation Study} \label{sec:sim_study}
We combined different mean functions and error processes to simulate the quality of an ML model over time. To generate the mean qualities, we used a constant function $\mu_1$, a monotonically decreasing function $\mu_2$, a non-monotonically decreasing function $\mu_3$ and a piecewise constant function $\mu_4$. The latter function $\mu_4$ clearly violates Assumption \ref{ass:main} (3) and is used to test the methods' robustness. More specifically, we chose
\begin{align*}
	\mu_1(t) &=0.9,\\
	\mu_2(t) &= \left\{\begin{array}{ll}
0.9&\text{for}~t\le \tfrac{1}{4}\\
0.8 + \tfrac{1}{10}\sin(2\pi t)&\text{for}~\tfrac{1}{4}<t\le \tfrac{3}{4}\\
0.7&\text{for}~\tfrac{3}{4}<t,
\end{array} \right. \\
	\mu_3(t)&=  \left\{\begin{array}{ll}
	 0.85 + \tfrac{1}{20} \sin(8 \pi t)&\text{for}~t \le \tfrac{1}{4}\\
	0.85 + \tfrac{1}{20} \sin(8 \pi t) - 0.145 (t- 1/4)&\text{for}~ t > \tfrac{1}{4},
	\end{array} \right. \\
	\mu_4(t)& =  \left\{\begin{array}{ll}
	0.9&\text{for}~t \le \tfrac{1}{5}\\
	0.7&\text{for}~ t > \tfrac{1}{5}.
	\end{array} \right.
\end{align*}

While the functions  $\mu_1, \mu_2$ and $\mu_4$ start at $90\%$, $\mu_3$ starts at $85\%$ for $t=0$. $\mu_1$ is constant and the other functions decrease to a minimum value of $70\%$. In particular, the null hypothesis is true for $\mu_1$ and any $\Delta \geq 0$, for $\mu_2$ and $\mu_4$, it is true with $\Delta \ge 20\%$ and for $\mu_3$ with $\Delta \ge 16\%$. 

Further, we used three different classes of error processes. Given a sequence of independent, standard normally distributed random variables $\{\eta_i\}_{i\in\Z}$, we defined
\begin{align*}
	\eps_i^{(\iid)} & = \tfrac{1}{20} \eta_i, \\
	\eps_i^{(\ma)} & = \tfrac{1}{20} \sqrt{\tfrac{4}{5}} \big(\eta_i + \tfrac{1}{2} \eta_{i-1}\big),\\
	\eps_i^{(\ar)} & = \tfrac{1}{20} \sqrt{\tfrac{15}{16}} \big(\eta_i + \tfrac{1}{4}\eps_{i-1}^{(\ar)} \big),
\end{align*}
for $i\in\Z$. In particular, all of the error processes have a variance of $\tfrac{1}{400}$, i.\,e., a standard deviation of $5\%$.

\begin{figure}
	\vskip -0.1in
	\begin{center}
		\includegraphics[width=0.45\textwidth]{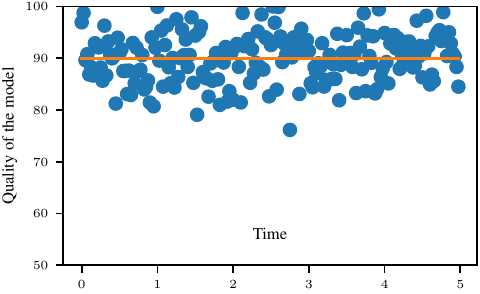} \hspace{0.1cm}
		\includegraphics[width=0.45\textwidth]{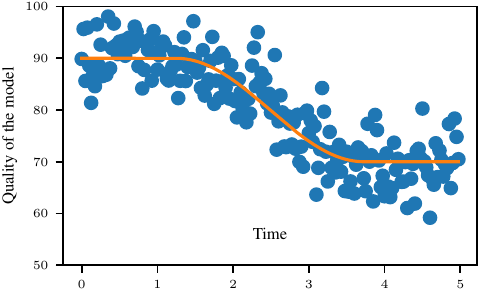} \hspace{0.1cm}
		\includegraphics[width=0.45\textwidth]{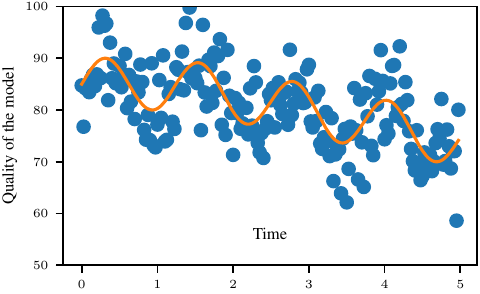} \hspace{0.1cm}
		\includegraphics[width=0.45\textwidth]{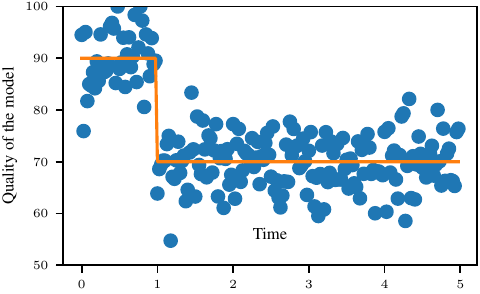} \hspace{0.1cm}		\vskip -0.1in
		\caption{Exemplary simulated model qualities for \iid~ errors and $n=40$. The model quality is given in \%. Top left: $\mu_1$. Top right: $\mu_2$. Bottom left: $\mu_3$. Bottom right: $\mu_4$.}
		\label{fig:sim_data_ex}
	\end{center}
\end{figure}

Based on these mean functions and error processes, we simulated model qualities given by 
\begin{equation*}
	X_t^{(i, j)} = \mu_i\big(\tfrac{t}{5n}\big) + \eps_t^{(j)}
\end{equation*}
for $t = 1, \dots, 5n,~ n = 40, 100, 200,~i\in\{1, 2, 3, 4\}$ and $j\in \{\iid, \ma, \ar\}$. Examples of the simulated model qualities are displayed in Figure \ref{fig:sim_data_ex}. It can be seen that the observed quality is rather noisy and the naive monitoring scheme, which does not take random fluctuations into account, will raise many false positive alarms.

For each combination of mean function and error process, we simulated 1000 trajectories of model qualities. For the local linear estimation used in the proposed monitoring schemes \eqref{eq:mon_scheme} and \eqref{eq:approx_mon_scheme}, we used the quartic kernel $K(x) = \frac{15}{16} (1-x^2)^2$ and selected the bandwidth $h_n$ between $n/4$ and $n/2$ via cross validation with 10 folds.
Further, we chose the block length of the long-run variance estimator as
$$ m_n = \max\bigg\{ \bigg\lfloor \sqrt{\tfrac{|\hat{\gamma}_1| + \dots + |\hat{\gamma}_4|}{|\hat{\gamma}_0| + \dots + |\hat{\gamma}_4|}} n^{1/3} \bigg\rfloor, 1 \bigg\}, $$
where $\hat{\gamma}_h$ denotes the empirical autocovariance of the residuals $\hat{\eps}_{i, n} = X_{i, n} - \tilde{\mu}(i/n)$ at lag $h$, for $h=0, \dots, 4$. Finally, we approximated the quantiles of the Gaussian approximation in \eqref{eq:approx_mon_scheme} based on 1000 simulated trajectories of the Gaussian process that is maximized by the supremum and chose the significance level $\alpha = 0.05$. 

\begin{figure}[h]
	\vskip -0.1in
	\begin{center}
		\includegraphics[width=0.45\textwidth]{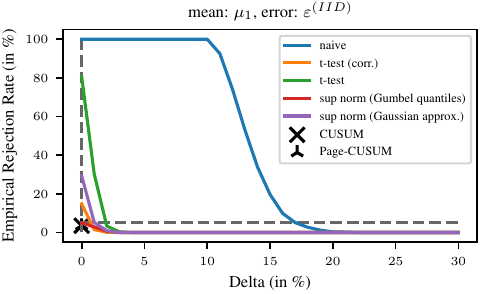} \hspace{0.1cm}
		\includegraphics[width=0.45\textwidth]{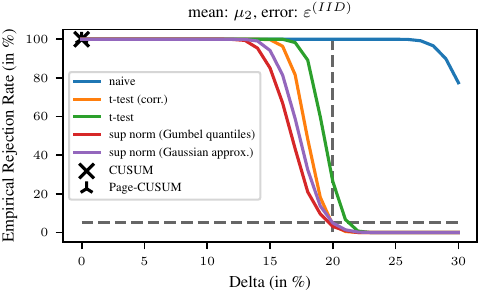}\\
		\includegraphics[width=0.45\textwidth]{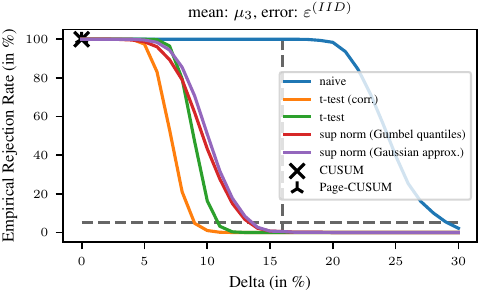}\hspace{0.1cm}
		\includegraphics[width=0.45\textwidth]{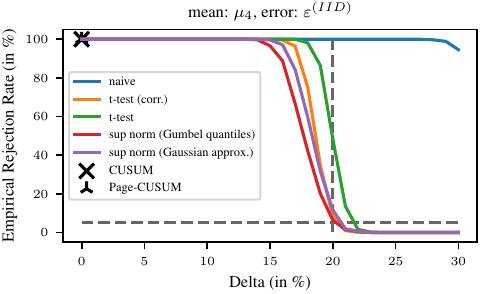}
		\vskip -0.1in
		\caption{Empirical rejection rates of the different monitoring schemes for $n=40$ and varying values of $\Delta$. The dashed horizontal line marks the level $\alpha=5\%$ and the dashed vertical line the boundary of the null hypothesis. Top left: $\mu_1$. Top right: $\mu_2$. Bottom left: $\mu_3$. Bottom right: $\mu_4$.}
		\label{fig:sim_data}
	\end{center}
\end{figure}

\begin{figure}[h]
	\vskip -0.1in
	\begin{center}
		\includegraphics[width=0.45\textwidth]{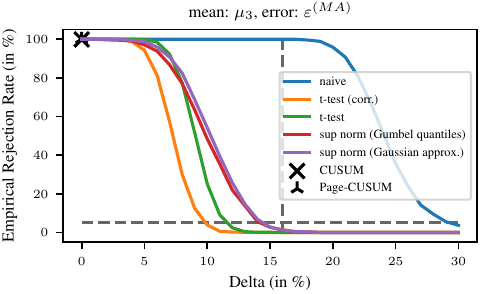}\hspace{0.1cm}
		\includegraphics[width=0.45\textwidth]{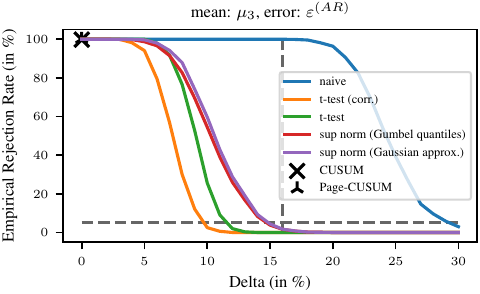}\\
		\includegraphics[width=0.45\textwidth]{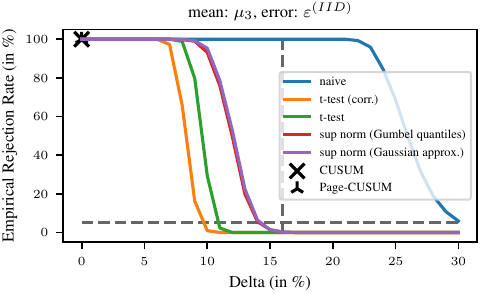}\hspace{0.1cm}
		\includegraphics[width=0.45\textwidth]{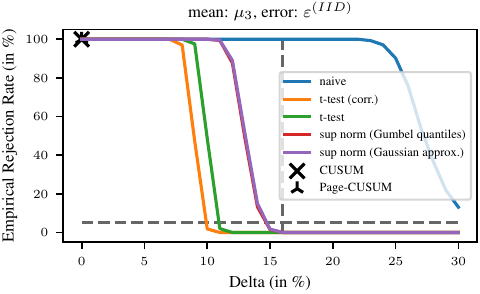}
		\vskip -0.1in
		\caption{Empirical rejection rates of the different monitoring schemes for  $\mu_3$ and varying values of $\Delta$. The dashed horizontal line marks the level $\alpha=5\%$ and the dashed vertical line the boundary of the null hypothesis. Top left: MA errors, $n=40$. Top right: AR errors, $n=40$. Bottom left: IID errors, $n=100$. Bottom right: IID errors, $n=200$.}
		\label{fig:sim_data_variations}
	\end{center}
\end{figure}

The results of the experiments are displayed in Figure \ref{fig:sim_data} for \iid~ errors and $n=40$ and in Figure \ref{fig:sim_data_variations} for other error types and larger values of $n$. More detailed results are provided in Tables \ref{tab:sim_data_iid}, \ref{tab:sim_data_cusum}, \ref{tab:sim_data_ma} and \ref{tab:sim_data_ar} of Appendix \ref{sec:app_sim_study}.

In the plots, the vertical dashed line marks the boundary of the null hypothesis of no relevant deviation. It can be easily seen, that the naive monitoring scheme \eqref{eq:naive_scheme} rejects the null hypothesis too often and leads to many false alarms. For the constant model quality $\mu_1$, only the proposed monitoring scheme \eqref{eq:mon_scheme} based on Gumbel quantiles and the two CUSUM-based monitoring schemes have empirical rejection rates below the level of 5\%. In contrast, the t-test based monitoring scheme \eqref{eq:t_test} without $\alpha$ level correction, falsely rejects the null hypothesis in about 80\% of the cases. For the model quality functions $\mu_2$ and $\mu_4$, the proposed monitoring schemes \eqref{eq:mon_scheme} and \eqref{eq:approx_mon_scheme} yield similar results as the benchmark \eqref{eq:t_test_corr}, while \eqref{eq:t_test} does not have level $\alpha=5\%$ at the boundary of the null hypothesis. For $\mu_3$ all non-naive monitoring schemes are overly conservative, but the proposed schemes \eqref{eq:mon_scheme} and \eqref{eq:approx_mon_scheme} are more powerful than the t-test based ones. For error processes with temporal dependence, the power of the latter monitoring schemes remains similar, while it increases for the proposed monitoring schemes (see top of Figure \ref{fig:sim_data_variations}). For larger values of $n$, the power of \eqref{eq:mon_scheme} and \eqref{eq:approx_mon_scheme} further increases, while the t-test based monitoring schemes do not detect relevant deviations close to the boundary of the null hypothesis.

For $\Delta = 0$, the CUSUM-based monitoring schemes \eqref{eq:cusum_test} and \eqref{eq:page_cusum_test} have empirical rejection rates below $5\%$ under the null hypothesis and find deviations reliably. However, they are only applicable for $\Delta = 0$ and in real applications, the tolerable deviation is often larger than 0.

In Tables \ref{tab:sim_data_dev} and \ref{tab:sim_data_cusum} of Appendix \ref{sec:app_sim_study}, the times of the first relevant deviation are displayed. Note that the estimated times are based only on cases, where a relevant deviation is detected. If the rejection rate is rather small, these estimates are based only on a few samples and have to be interpreted carefully. For the naive monitoring scheme \eqref{eq:naive_scheme}, the first relevant deviations are detected rather early, which is consistent with the high empirical rejection rates. Thus, the naive monitoring scheme leads to many, early (false) alarms. For the remaining monitoring schemes, the results are similar for $\mu_1$ and $\mu_2$. Only the t-test based monitoring scheme \eqref{eq:t_test} without $\alpha$ level correction yields slightly better estimates for $\mu_2$. Conversely, the estimated times of the proposed monitoring schemes \eqref{eq:mon_scheme} and \eqref{eq:approx_mon_scheme} are better for $\mu_3$ and $\mu_4$, especially for larger values of $n$ and smaller values of $\Delta$. 

In summary, a CUSUM-based monitoring scheme might be most suitable for $\Delta = 0$, while for $\Delta > 0$ the monitoring scheme \eqref{eq:mon_scheme} based on Gumbel quantiles seems to be most reliable as it has good power under the alternative and approximately level $\alpha$, which means that it detects relevant deviations but simultaneously does not yield too many false alarms.

\subsection{Real Data Experiments} \label{sec:real_data_exp}

When using ML models in real-world systems, in some cases we get feedback on how accurate the model's predictions are, while in others we do not receive any feedback. In the first case, we can monitor quality metrics, such as the model's accuracy, while in the latter case we can only use quantities directly linked to predictions, such as the confidence. For the following experiments, we used both accuracies and confidences to evaluate the proposed method in situations with and without feedback.

We used two models and three different data regimes to generate observations of model qualities over time. More specifically, we used a logistic regression classifier (\LogReg) and a multilayer perceptron (\NN) with three hidden layers consisting of 10 neurons each. For the \NN, we used two different training strategies. First, we trained an initial model with fixed weights (\NNI) and second, we trained an \NN~ and continuously re-trained it at every time step (\NNC).

We used the classifiers to separate two classes of noisy data. More specifically, we generated an initial training data set of 1000 samples and subsequently 2000 epochs of 100 samples each. We measured the models' accuracies and confidences at every epoch, yielding a life span of 2000. The neural networks were trained with batches of size 10.

\begin{figure}[h]
	\vskip -0.1in
	\begin{center}
		\includegraphics{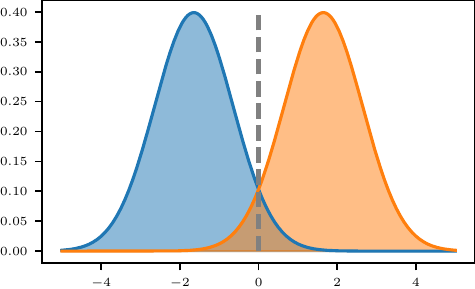}
		\vskip -0.1in
		\caption{Probability density functions of two normal distributions with variance 1 and mean $-1.645$ and $1.645$, such that $\pr(X_1 \le 0) = \pr(X_2 \ge 0) = 95\%$.}
		\label{fig:norm_data}
	\end{center}
\end{figure}

Given a probability $p\in(0, 1)$, we generated samples for the different classes such that an optimal classifier would have an (expected) accuracy of $1-p$. More specifically, we drew from two normal distributions with unit variance such that $\pr(X_1 \le 0) = \pr(X_2 \ge 0) = 1-p$, where $X_1$ is a random variable from the first and $X_2$ from the second distribution. This property holds if $X_1 \sim \Nc(-\delta, 1)$ and $X_2\sim \Nc(\delta, 1)$ for $\delta = \Phi^{-1}(1- p)$, where $\Phi$ denotes the cumulative distribution function of the standard normal distribution. A classifier distinguishing between these two classes will have an (expected) accuracy of $1 - p$, if both classes are equally likely, since $p$ of the samples from each class will fall into a region where the other class predominates, see Figure \ref{fig:norm_data}.

First, we generate \textit{stable} data, where $p = 0.05$ is constant over all epochs and the proportion of both classes is 50\%. Second, we generate data with a \textit{concept drift}, where again $p= 0.05$ is constant over the entire time span, but the proportion of the first class changes over time. More specifically, its proportion is constant at 50\% during the first 400 epochs, it increases linearly to 90\% during the next 800 epochs and remains constant at 90\% for the last 800 epochs. Finally, we generate data with a \textit{data drift}, where the proportion of both classes is constant at 50\% and the probability $p$ changes over time. As before, it remains constant at 0.05 for the first 400 epochs, increases linearly to 0.1 for the next 800 epochs and remains constant at 0.1 for the final 800 epochs.

We set $n=100$ and monitored deviations from epoch 101 until the end of the models' life spans. This is equivalent to a scenario, where a model is initially trained with a training data set, then deployed but only used as a shadow model for 100 time steps, i.\,e., the model's predictions are used for analysis of its behavior and not to control the system, and finally used to make predictions that control the system for another 1900 time steps. 

\begin{figure}
	\vskip -0.1in
	\begin{center}
		\includegraphics[width=0.45\textwidth]{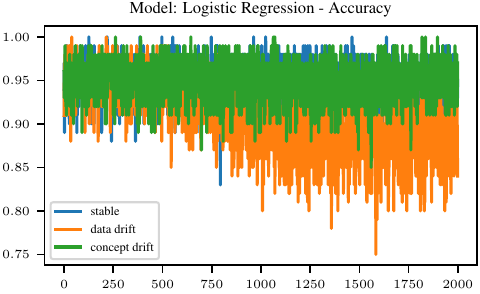} \hspace{0.1cm}
		\includegraphics[width=0.45\textwidth]{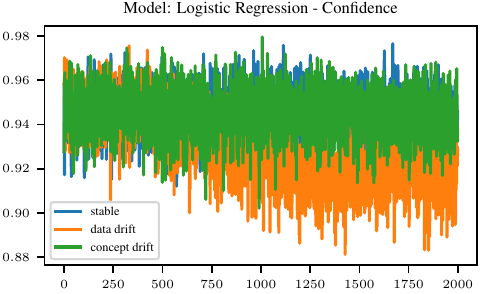} \hspace{0.1cm}
		\includegraphics[width=0.45\textwidth]{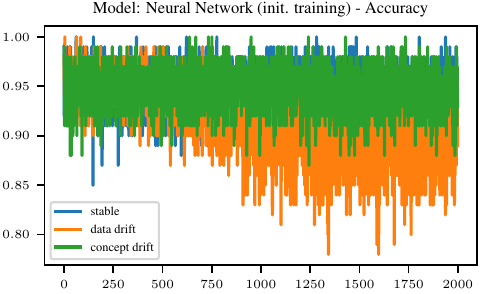} \hspace{0.1cm}
		\includegraphics[width=0.45\textwidth]{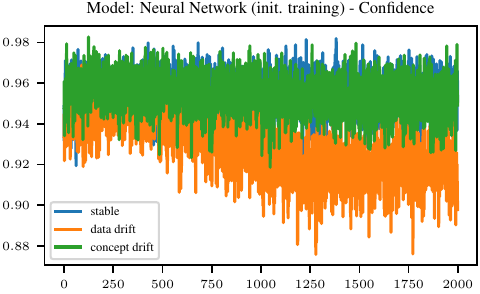} \hspace{0.1cm}		\includegraphics[width=0.45\textwidth]{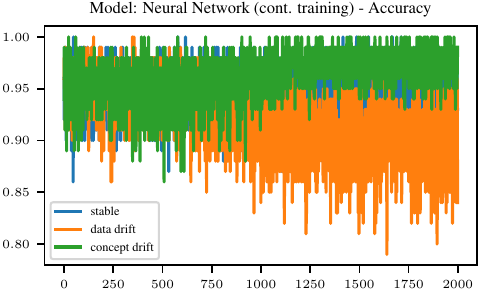} \hspace{0.1cm}
		\includegraphics[width=0.45\textwidth]{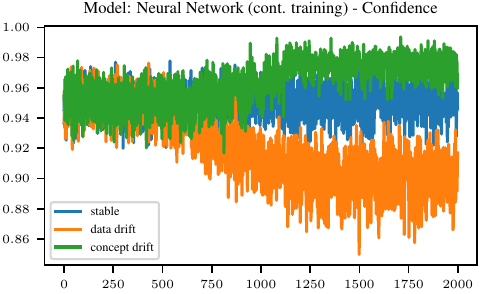} \hspace{0.1cm}
		\vskip -0.1in
		\caption{Exemplary histories of the model quality for different data regimes. The plots on the left are based on the accuracy, while the plots on the right are based on the confidence. Top: logistic regression (\LogReg). Center: neural network with initial training only (\NNI). Bottom: neural network with continuous training (\NNC).}
		\label{fig:real_data_ex}
	\end{center}
\end{figure}

Examples of the resulting model qualities are displayed in Figure \ref{fig:real_data_ex}. The observed accuracies and confidences of \LogReg~ and \NNI~ seem rather similar with robust values for stable data and data with concept drift. Only in the case of a data drift, a structural deviation can be seen for all models. It is specifically visible after 1000 epochs and for \LogReg~ and \NNI~ more pronounced in the accuracies. For the \NNC, a structural deviation can be seen in the data drift too, but here it is stronger in the observed confidences. Finally, in case of data with a concept drift, the continuously training \NNC~ slightly improved over time with respect to the accuracies and the confidences. Thus, it can be expected to find relevant deviations for all models based on data with data drift and for the \NNC~ based on data with concept drift.

As before, we chose the significance level $\alpha = 0.05$ and used the same specifications for the proposed monitoring schemes \eqref{eq:mon_scheme} and \eqref{eq:approx_mon_scheme} as described in Section \ref{sec:sim_study}.

\begin{figure}[h]
	\vskip -0.1in
	\begin{center}
		\includegraphics[width=0.45\textwidth]{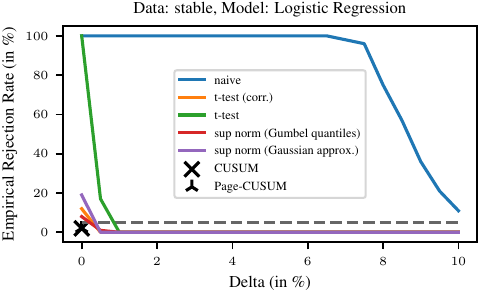} \hspace{0.1cm}
		\includegraphics[width=0.45\textwidth]{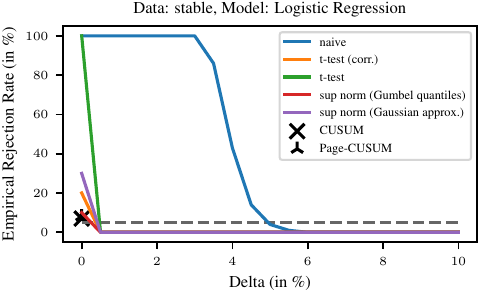}\\
		\includegraphics[width=0.45\textwidth]{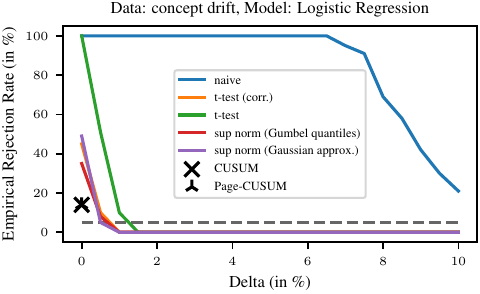} \hspace{0.1cm}
		\includegraphics[width=0.45\textwidth]{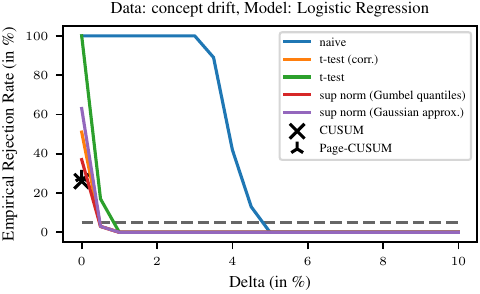} \\
		\includegraphics[width=0.45\textwidth]{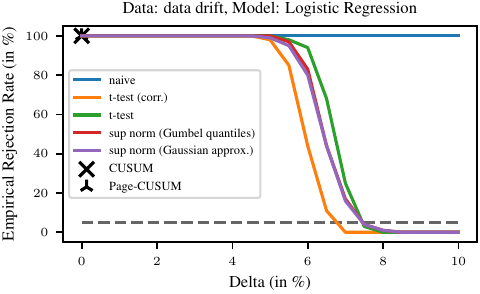} \hspace{0.1cm}
		\includegraphics[width=0.45\textwidth]{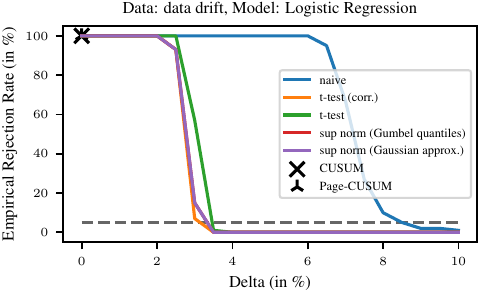} \\
		\vskip -0.1in
		\caption{Empirical rejection rates of the different monitoring schemes for a logistic regression and varying values of $\Delta$ based on the models' \textit{accuracies} (left) and \textit{confidences} (right) for stable data (top), data with concept drift (center) and data with data drift (bottom). The dashed horizontal line marks the level $\alpha=5\%$.}
		\label{fig:real_data_log_reg}
	\end{center}
\end{figure}

The results of the experiments for the \LogReg~ model based on the accuracies and confidences are displayed in 
Figure \ref{fig:real_data_log_reg}. Moreover, results for the \NNI~ and \NNC~ models are displayed in Figures \ref{fig:real_data_nn} and \ref{fig:real_data_nn_conf} of Appendix \ref{sec:app_real_data}. More detailed results are provided in Tables \ref{tab:real_data_cusum} and \ref{tab:real_data}.

In those cases with a constant accuracy, i.\,e., for all models with stable data and the models \LogReg~ and \NNI~ with data with concept drift, the CUSUM-based monitoring schemes and \eqref{eq:mon_scheme} yield the best results as they (approximately) have the specified level $\alpha=5\%$. The empirical rejection rates of the monitoring scheme based on approximated quantiles \eqref{eq:approx_mon_scheme} and the t-test based monitoring scheme with $\alpha$ level correction \eqref{eq:t_test_corr} slightly exceed the level of the test, whereas the scheme without $\alpha$ level correction \eqref{eq:t_test} drastically exceeds $\alpha=5\%$. Again, the naive monitoring scheme \eqref{eq:naive_scheme} rejects the null hypothesis too often, yielding many false alarms.

In those cases with a non-constant accuracy, the empirical rejection rates of \eqref{eq:mon_scheme} and \eqref{eq:approx_mon_scheme} are between those of the t-test based monitoring schemes, where \eqref{eq:t_test_corr} seems to be slightly conservative and \eqref{eq:t_test} seems to exceed the predefined level close to the boundary of the null hypothesis. A notable exception is the model \NNC~ with data with concept drift, as in this case the schemes \eqref{eq:mon_scheme} and \eqref{eq:approx_mon_scheme} seem to be slightly more conservative than \eqref{eq:t_test_corr}. Note that straight forward calculations show that the expected accuracy in this case is approximately 97.42\%, so $\Delta\approx 2.4\%$.

For $\Delta=0$, the CUSUM-based monitoring schemes seem to yield the best results. When the confidences of the models' predictions are monitored instead of their accuracies, similar results can be observed, although the differences between the monitoring schemes seem to vanish.

Thus, as the results of the simulated model histories suggested, the CUSUM-based monitoring schemes seem to be most suitable for $\Delta = 0$, while for $\Delta > 0$ the monitoring schemes \eqref{eq:mon_scheme} based on Gumbel quantiles and \eqref{eq:t_test_corr} based on the t-test with $\alpha$ level correction seem to be most reliable as they have good power under the alternative, yet have approximately level $\alpha$ under the null hypothesis.

\section{Conclusion}  \label{sec:con}

In this work, we proposed two monitoring schemes for ML models that are able to detect relevant deviations of the models' qualities. In particular, we proved that their probability of false alarms can be bounded asymptotically by some predefined level $\alpha\in(0, 1)$ and that they are consistent.

Further, we compared the proposed methods to alternative monitoring schemes for relevant hypotheses. For the simulated data experiments, the proposed monitoring schemes, in particular the one based on Gumbel quantiles, seem to outperform the alternatives. For the real data experiments, the t-test with corrected level seems to match the performance of the proposed methods. Generally, for $\Delta = 0$, the CUSUM-based monitoring schemes seem preferable. 

From a theoretic point of view however, the proposed monitoring schemes are preferable over the alternatives, as they require weaker assumptions. Throughout this work, the error process $\{\eps_i\}_{i\in\Z}$ was assumed to be stationary in order to ensure that the long-run variance is constant over time. Similar results hold for non-stationary error processes as long as the long-run variance changes smoothly over time.

In the proposed monitoring schemes, we gave the same weight to all time points. This limited us to a finite time horizon, as it would be impossible to bound the probability of false alarms otherwise. When moving from this \textit{closed-end} to an \textit{open-end} approach, we need to assign decreasing weights to time points in the future, which ultimately reduces the power of the monitoring scheme in the farther future. In practice, however, the benefit of an open-end monitoring scheme might be preferable to a scheme with finite time horizon. Therefore it would be fruitful to extend the proposed monitoring schemes to indefinite time horizons.

As for classic hypotheses, i.\,e., for $\Delta = 0$, CUSUM-based monitoring schemes seem preferable, it might be worthwhile to extend these schemes to relevant hypotheses for $\Delta > 0$. Further, in many applications we often want to monitor multiple quantities. Thus, it would be useful to have similar results for multivariate quantities that consider dependencies between the coordinates. Finally, in the field of machine learning many milestones are due to high-quality open data sets. Thus, a standardized benchmark data set with ML model histories and labeled failures would help drive research in the field of ML model monitoring and make results comparable. 

\section*{Acknowledgment}

This work is supported by the Ministry of Economics, Innovation, Digitization and Energy of the State of North Rhine-Westphalia and the European Union, grant IT-2-2-023 (VAFES).

\bibliography{bibliography}

\newpage
\appendix

\section{Proof of Theorem \ref{thm:main}}
\label{sec:app_theory}

\begin{proof}
	Let $\|f\|_\infty = \sup_{t\in[0,T]}|f(t)|$ denote the sup-norm of a function $f:[0, T]\to\R$. With this notation, the maximal deviation of $\mu$ from the baseline $g(\mu)$ can be written as $\|\mu - g(\mu)\|_\infty$, and the corresponding estimator as $\|\tilde{\mu}_{h_n} - \hat{g}_n\|_\infty$. In particular, $\|\mu - g(\mu)\|_\infty \le \Delta$ under the joint null hypothesis  $\bar{H}_0^{(T)}$. Further, let $\convw$ denote weak convergence of random variables.

	By Theorem 3.1 of \cite{bucher2021}, there exists a random variable $G$ such that	
	$$ \frac{\sqrt{nh_n}\ell_n}{\hat{\sigma}_{lrv}\|K^*\|_2} \big( \|\tilde{\mu}_{h_n} - \hat{g}_n\|_\infty - \|\mu - g(\mu)\|_\infty \big) - \ell_n^2 \convw G, $$ 
	where $\Gum_0((-\infty, x]) \le \pr(G \le x)$, if $\|\mu - g(\mu)\|_\infty > 0$, and $\Gum_{\log(2)}((-\infty, x]) \le \pr(G \le x)$, if $\|\mu - g(\mu)\|_\infty = 0$. 
	
	Under the null hypothesis $\bar{H}_0^{(T)}$, for $\Delta > 0$, we obtain
	\begin{align*}
	&\pr\bigg(\big|\tilde{\mu}_{h_n}(t)-\hat{g}_n\big| > \Delta + \big(q_{0,1-\alpha}+\ell_n^2\big) \frac{\hat{\sigma}_{lrv}\|K^*\|_2}{\sqrt{nh_n}\ell_n}~\text{for some}~t\in[0, T]\bigg)\\
	& =  \pr\bigg(\|\tilde{\mu}_{h_n}-\hat{g}_n\|_\infty > \Delta + \big(q_{0,1-\alpha}+\ell_n^2\big) \frac{\hat{\sigma}_{lrv}\|K^*\|_2}{\sqrt{nh_n}\ell_n}\bigg)\\
	& \le  \pr\bigg(\|\tilde{\mu}_{h_n}-\hat{g}_n\|_\infty > \|\mu - g(\mu)\|_\infty + \big(q_{0,1-\alpha}+\ell_n^2\big) \frac{\hat{\sigma}_{lrv}\|K^*\|_2}{\sqrt{nh_n}\ell_n}\bigg)\\
	& =  \pr\bigg(\frac{\sqrt{nh_n}\ell_n}{\hat{\sigma}_{lrv}\|K^*\|_2} \big( \|\tilde{\mu}_{h_n} - \hat{g}_n\|_\infty - \|\mu - g(\mu)\|_\infty \big) - \ell_n^2 > q_{0,1-\alpha}\bigg),
	\end{align*}
	where the right-hand side converges to $\pr(G > q_{0,1-\alpha})$, which can be further bound by
	$$\pr(G > q_{0,1-\alpha}) = 1 - \pr(G \le q_{0,1-\alpha}) \le 1 - \Gum_0((-\infty, q_{0,1-\alpha}]) = \alpha.$$
	Analogously, we obtain the claimed convergence for $\Delta = 0$ when $q_{0,1-\alpha}$ is replaced by $q_{\log(2),1-\alpha}$.
	
	With similar calculations, it holds that 
	\begin{align}\label{eq:consistency}
	\pr\bigg(&\big|\tilde{\mu}_{h_n}(t)-\hat{g}_n\big| > \Delta + \big(q_{0,1-\alpha}+\ell_n^2\big) \frac{\hat{\sigma}_{lrv}\|K^*\|_2}{\sqrt{nh_n}\ell_n}~\text{for some}~t\in[0, T]\bigg) \notag\\
	= \pr\bigg(&\frac{\sqrt{nh_n}\ell_n}{\hat{\sigma}_{lrv}\|K^*\|_2} \big( \|\tilde{\mu}_{h_n} - \hat{g}_n\|_\infty - \|\mu - g(\mu)\|_\infty \big) - \ell_n^2\\
	& > q_{0,1-\alpha} - \frac{\sqrt{nh_n}\ell_n}{\hat{\sigma}_{lrv}\|K^*\|_2} \big( \|\mu - g(\mu)\|_\infty - \Delta \big)\bigg). \notag
	\end{align}
	From the proof of Corollary 4.3 of \cite{bucher2021}, it follows that $\|\tilde{\mu}_{h_n} - \hat{g}_n\|_\infty - \|\mu - g(\mu)\|_\infty$ converges to zero in probability, i.\,e., $\|\tilde{\mu}_{h_n} - \hat{g}_n\|_\infty - \|\mu - g(\mu)\|_\infty = o_\pr(1)$. Hence, the left-hand side in the latter probability is of order $o_\pr(\sqrt{nh_n}\ell_n)$, while the right-hand side dominates the asymptotic behavior for $\|\mu - g(\mu)\|_\infty \neq \Delta$. In particular, $\|\mu - g(\mu)\|_\infty > \Delta$ under the alternative $\bar{H}_1^{(T)}$ such that the probability in \eqref{eq:consistency} converges to 1 as $n\to\infty$.
\end{proof}

\section{Additional Empirical Results}

\subsection{Additional Results of the Simulation Study}
\label{sec:app_sim_study}

\begin{table}[h]
	\footnotesize
	\begin{center}
	\begin{tabular}{l| rrr | rrr | rrr | rrr | rrr }	\hline \hline
	&\multicolumn{3}{c|}{mon. scheme \eqref{eq:naive_scheme} }&\multicolumn{3}{c|}{mon. scheme \eqref{eq:t_test}} & \multicolumn{3}{c|}{mon. scheme \eqref{eq:t_test_corr}}&\multicolumn{3}{c|}{mon. scheme \eqref{eq:mon_scheme}}&\multicolumn{3}{c}{mon. scheme \eqref{eq:approx_mon_scheme}}\\
	$\Delta$ & 40 & 100 & 200 & 40 & 100 & 200 & 40 & 100 & 200 & 40 & 100 & 200 & 40 & 100 & 200 \\ 
	\hline 
	\addlinespace[.2cm]
	\multicolumn{16}{l}{\quad\textit{Panel A: $\mu_1$}} \\ 
	\textbf{0} & \textbf{100.0} & \textbf{100.0} & \textbf{100.0} & \textbf{80.6} & \textbf{79.5} & \textbf{80.2} & \textbf{14.8} & \textbf{9.8} & \textbf{7.7} & \textbf{4.9} & \textbf{2.4} & \textbf{0.8} & \textbf{29.3} & \textbf{25.0} & \textbf{21.2}\\
	1 & 100.0 & 100.0 & 100.0 & 30.2 & 11.1 & 2.9 & 1.5 & 0.2 & 0.0 & 2.9 & 0.8 & 0.0 & 5.2 & 1.6 & 0.1\\
	2 & 100.0 & 100.0 & 100.0 & 3.3 & 0.2 & 0.0 & 0.1 & 0.0 & 0.0 & 0.6 & 0.1 & 0.0 & 0.9 & 0.1 & 0.0\\
	10 & 100.0 & 100.0 & 100.0 & 0.0 & 0.0 & 0.0 & 0.0 & 0.0 & 0.0 & 0.0 & 0.0 & 0.0 & 0.0 & 0.0 & 0.0\\
	15 & 19.7 & 41.8 & 65.3 & 0.0 & 0.0 & 0.0 & 0.0 & 0.0 & 0.0 & 0.0 & 0.0 & 0.0 & 0.0 & 0.0 & 0.0\\
	20 & 0.3 & 1.1 & 2.9 & 0.0 & 0.0 & 0.0 & 0.0 & 0.0 & 0.0 & 0.0 & 0.0 & 0.0 & 0.0 & 0.0 & 0.0\\
	\addlinespace[.2cm]
	\multicolumn{16}{l}{\quad\textit{Panel B: $\mu_2$}} \\ 
	12 & 100.0 & 100.0 & 100.0 & 100.0 & 100.0 & 100.0 & 100.0 & 100.0 & 100.0 & 99.9 & 100.0 & 100.0 & 100.0 & 100.0 & 100.0\\
	14 & 100.0 & 100.0 & 100.0 & 100.0 & 100.0 & 100.0 & 100.0 & 100.0 & 100.0 & 95.4 & 100.0 & 100.0 & 99.1 & 100.0 & 100.0\\
	16 & 100.0 & 100.0 & 100.0 & 100.0 & 100.0 & 100.0 & 96.4 & 100.0 & 100.0 & 67.0 & 98.3 & 100.0 & 81.5 & 99.5 & 100.0\\
	18 & 100.0 & 100.0 & 100.0 & 89.2 & 98.2 & 99.9 & 48.3 & 78.1 & 96.6 & 21.0 & 43.4 & 76.2 & 32.5 & 58.5 & 90.1\\
	19 & 100.0 & 100.0 & 100.0 & 59.9 & 78.2 & 91.5 & 18.2 & 23.3 & 42.5 & 9.6 & 9.7 & 16.6 & 13.7 & 18.4 & 32.3\\
	\textbf{20} & \textbf{100.0} & \textbf{100.0} & \textbf{100.0} & \textbf{26.3} & \textbf{23.7} & \textbf{24.9} & \textbf{3.8} & \textbf{1.0} & \textbf{1.0} & \textbf{3.1} & \textbf{0.7} & \textbf{0.7} & \textbf{4.9} & \textbf{2.4} & \textbf{2.1}\\
	21 & 100.0 & 100.0 & 100.0 & 6.7 & 0.8 & 0.1 & 0.5 & 0.0 & 0.0 & 0.9 & 0.1 & 0.1 & 1.4 & 0.1 & 0.1\\
	\addlinespace[.2cm]
	\multicolumn{16}{l}{\quad\textit{Panel C: $\mu_3$}} \\ 
	6 & 100.0 & 100.0 & 100.0 & 99.8 & 100.0 & 100.0 & 83.1 & 100.0 & 100.0 & 96.1 & 100.0 & 100.0 & 98.4 & 100.0 & 100.0\\
	8 & 100.0 & 100.0 & 100.0 & 79.5 & 98.9 & 100.0 & 21.0 & 65.9 & 97.0 & 79.0 & 99.8 & 100.0 & 85.5 & 99.9 & 100.0\\
	10 & 100.0 & 100.0 & 100.0 & 16.4 & 29.4 & 48.6 & 1.0 & 0.9 & 1.8 & 43.3 & 93.2 & 100.0 & 50.9 & 95.4 & 100.0\\
	12 & 100.0 & 100.0 & 100.0 & 0.3 & 0.0 & 0.0 & 0.0 & 0.0 & 0.0 & 15.0 & 49.5 & 87.7 & 17.7 & 52.8 & 89.1\\
	14 & 100.0 & 100.0 & 100.0 & 0.0 & 0.0 & 0.0 & 0.0 & 0.0 & 0.0 & 2.2 & 5.6 & 13.2 & 2.7 & 6.4 & 15.0\\
	15 & 100.0 & 100.0 & 100.0 & 0.0 & 0.0 & 0.0 & 0.0 & 0.0 & 0.0 & 0.7 & 1.4 & 1.5 & 0.7 & 1.4 & 1.7\\
	\textbf{16} & \textbf{100.0} & \textbf{100.0} & \textbf{100.0} & \textbf{0.0} & \textbf{0.0} & \textbf{0.0} & \textbf{0.0} & \textbf{0.0} & \textbf{0.0} & \textbf{0.5} & \textbf{0.2} & \textbf{0.0} & \textbf{0.5} & \textbf{0.2} & \textbf{0.0}\\
	\addlinespace[.2cm]
	\multicolumn{16}{l}{\quad\textit{Panel D: $\mu_4$}} \\ 
	14 & 100.0 & 100.0 & 100.0 & 100.0 & 100.0 & 100.0 & 100.0 & 100.0 & 100.0 & 99.7 & 100.0 & 100.0 & 100.0 & 100.0 & 100.0\\
	16 & 100.0 & 100.0 & 100.0 & 100.0 & 100.0 & 100.0 & 99.6 & 100.0 & 100.0 & 88.8 & 99.9 & 100.0 & 97.1 & 100.0 & 100.0\\
	18 & 100.0 & 100.0 & 100.0 & 98.1 & 100.0 & 100.0 & 75.3 & 94.5 & 100.0 & 42.2 & 75.4 & 97.1 & 59.7 & 89.1 & 99.3\\
	19 & 100.0 & 100.0 & 100.0 & 86.5 & 94.7 & 99.5 & 34.9 & 51.8 & 68.0 & 20.2 & 33.6 & 55.5 & 31.7 & 51.0 & 73.7\\
	\textbf{20} & \textbf{100.0} & \textbf{100.0} & \textbf{100.0} & \textbf{48.4} & \textbf{49.9} & \textbf{46.6} & \textbf{8.3} & \textbf{5.3} & \textbf{3.2} & \textbf{6.3} & \textbf{5.8} & \textbf{7.6} & \textbf{11.6} & \textbf{11.1} & \textbf{14.9}\\
	21 & 100.0 & 100.0 & 100.0 & 13.6 & 4.9 & 0.7 & 1.0 & 0.0 & 0.0 & 1.5 & 0.5 & 0.4 & 1.9 & 1.0 & 0.9\\
	22 & 100.0 & 100.0 & 100.0 & 1.6 & 0.0 & 0.0 & 0.1 & 0.0 & 0.0 & 0.5 & 0.1 & 0.0 & 0.7 & 0.1 & 0.0\\
	\hline \hline
\end{tabular}
	\end{center}
	\medskip
	\caption{Empirical rejection rates of various monitoring schemes for the hypotheses  \eqref{eq:null}, different mean functions $\mu$, IID errors and sample sizes $n=40, 100, 200$. }
	\label{tab:sim_data_iid}
\end{table}

\begin{table}
	\small
	\begin{center}
	\begin{tabular}{l| rrr | rrr | rrr | rrr | rrr | r}	\hline \hline
	&\multicolumn{3}{c|}{mon. scheme \eqref{eq:naive_scheme} }&\multicolumn{3}{c|}{mon. scheme \eqref{eq:t_test}} & \multicolumn{3}{c|}{mon. scheme \eqref{eq:t_test_corr}}&\multicolumn{3}{c|}{mon. scheme \eqref{eq:mon_scheme}}&\multicolumn{3}{c|}{mon. scheme \eqref{eq:approx_mon_scheme}}&first \\
	$\Delta$ & 40 & 100 & 200 & 40 & 100 & 200 & 40 & 100 & 200 & 40 & 100 & 200 & 40 & 100 & 200 & dev.\\ 
	\hline 
	
	\addlinespace[.2cm]
	\multicolumn{17}{l}{\quad\textit{Panel A: $\mu_1$,  IID errors}} \\
	0 & 1.00 & 1.00 & 1.00 & 2.57 & 2.54 & 2.59 & 3.06 & 3.14 & 3.18 & 2.73 & 3.18 & 2.51 & 2.72 & 2.66 & 2.75 & -\\
	1 & 1.00 & 1.00 & 1.00 & 3.02 & 3.13 & 2.96 & 3.74 & 2.28 & - & 2.59 & 3.17 & - & 2.86 & 3.10 & 2.96 & -\\
	2 & 1.01 & 1.00 & 1.00 & 3.43 & 2.35 & - & 2.85 & - & - & 2.72 & 4.73 & - & 2.39 & 4.72 & - & -\\
	10 & 1.63 & 1.27 & 1.14 & - & - & - & - & - & - & - & - & - & - & - & - & -\\
	15 & 2.91 & 2.78 & 2.62 & - & - & - & - & - & - & - & - & - & - & - & - & -\\
	20 & 2.24 & 3.09 & 2.83 & - & - & - & - & - & - & - & - & - & - & - & - & -\\

	\addlinespace[.2cm]
	\multicolumn{17}{l}{\quad\textit{Panel B: $\mu_2$,  IID errors}} \\
	12 & 1.91 & 1.64 & 1.46 & 3.33 & 3.27 & 3.24 & 3.48 & 3.38 & 3.32 & 3.59 & 3.40 & 3.32 & 3.53 & 3.37 & 3.29 & 2.66\\
	14 & 2.17 & 1.94 & 1.78 & 3.52 & 3.46 & 3.42 & 3.70 & 3.57 & 3.51 & 3.87 & 3.62 & 3.50 & 3.78 & 3.57 & 3.47 & 2.83\\
	16 & 2.38 & 2.20 & 2.06 & 3.76 & 3.68 & 3.63 & 4.00 & 3.83 & 3.74 & 4.13 & 3.95 & 3.75 & 4.07 & 3.86 & 3.70 & 3.01\\
	18 & 2.56 & 2.40 & 2.27 & 4.11 & 4.02 & 3.94 & 4.34 & 4.29 & 4.15 & 4.26 & 4.33 & 4.23 & 4.28 & 4.27 & 4.16 & 3.24\\
	19 & 2.65 & 2.49 & 2.38 & 4.28 & 4.28 & 4.23 & 4.45 & 4.46 & 4.47 & 4.35 & 4.37 & 4.42 & 4.32 & 4.38 & 4.40 & 3.39\\
	20 & 2.74 & 2.58 & 2.46 & 4.40 & 4.47 & 4.52 & 4.51 & 4.53 & 4.69 & 4.35 & 4.64 & 4.57 & 4.41 & 4.53 & 4.45 & -\\
	21 & 2.82 & 2.67 & 2.55 & 4.50 & 4.50 & 4.53 & 4.58 & - & - & 4.32 & 4.54 & 4.80 & 4.35 & 4.54 & 4.80 & -\\

	\addlinespace[.2cm]
	\multicolumn{17}{l}{\quad\textit{Panel C: $\mu_3$,  IID errors}} \\
	6 & 1.04 & 1.01 & 1.01 & 3.93 & 3.77 & 3.71 & 4.45 & 4.13 & 3.87 & 3.84 & 3.33 & 3.02 & 3.76 & 3.29 & 2.98 & 2.01\\
	8 & 1.09 & 1.03 & 1.01 & 4.60 & 4.59 & 4.51 & 4.83 & 4.90 & 4.88 & 4.40 & 3.96 & 3.62 & 4.33 & 3.92 & 3.61 & 2.13\\
	10 & 1.21 & 1.06 & 1.03 & 4.89 & 4.95 & 4.95 & 4.94 & 4.95 & 4.97 & 4.73 & 4.65 & 4.47 & 4.68 & 4.63 & 4.43 & 3.28\\
	12 & 1.47 & 1.19 & 1.08 & 4.94 & - & - & - & - & - & 4.84 & 4.88 & 4.85 & 4.84 & 4.87 & 4.85 & 3.43\\
	14 & 1.90 & 1.51 & 1.27 & - & - & - & - & - & - & 4.89 & 4.91 & 4.91 & 4.90 & 4.91 & 4.91 & 4.54\\
	15 & 2.20 & 1.80 & 1.54 & - & - & - & - & - & - & 4.92 & 4.91 & 4.92 & 4.92 & 4.90 & 4.92 & 4.61\\
	16 & 2.49 & 2.02 & 1.71 & - & - & - & - & - & - & 4.94 & 4.93 & - & 4.94 & 4.92 & - & -\\

	\addlinespace[.2cm]
	\multicolumn{17}{l}{\quad\textit{Panel D: $\mu_4$,  IID errors}} \\
	14 & 1.00 & 1.00 & 1.00 & 1.84 & 1.79 & 1.76 & 1.92 & 1.86 & 1.83 & 1.69 & 1.55 & 1.54 & 1.61 & 1.55 & 1.53 & 1.00\\
	16 & 1.01 & 1.00 & 1.00 & 1.92 & 1.88 & 1.85 & 2.04 & 1.94 & 1.91 & 2.06 & 1.63 & 1.57 & 1.93 & 1.60 & 1.57 & 1.00\\
	18 & 1.01 & 1.01 & 1.00 & 2.17 & 1.98 & 1.94 & 2.62 & 2.30 & 2.03 & 2.42 & 2.15 & 1.78 & 2.39 & 2.04 & 1.72 & 1.00\\
	19 & 1.02 & 1.01 & 1.00 & 2.50 & 2.31 & 2.09 & 3.01 & 2.84 & 2.70 & 2.79 & 2.40 & 2.12 & 2.58 & 2.40 & 2.09 & 1.00\\
	20 & 1.03 & 1.01 & 1.01 & 2.89 & 2.83 & 2.86 & 3.36 & 3.25 & 3.16 & 3.02 & 2.39 & 2.00 & 2.90 & 2.29 & 2.02 & -\\
	21 & 1.04 & 1.01 & 1.01 & 3.19 & 3.32 & 3.87 & 2.99 & - & - & 3.15 & 2.32 & 1.58 & 3.21 & 2.37 & 1.93 & -\\
	22 & 1.05 & 1.02 & 1.01 & 3.20 & - & - & 2.62 & - & - & 3.44 & 1.46 & - & 3.70 & 1.46 & - & -\\

	\hline \hline
\end{tabular}
	\end{center}
	\medskip
	\caption{Estimator for the first deviation of various monitoring schemes for the hypotheses  \eqref{eq:null}, different mean functions $\mu$, IID errors and sample sizes $n=40, 100, 200$. Out of the 1000 simulated trajectories, only those are used to estimate the time of the first relevant deviation, for which the null hypothesis was rejected at some point.}
	\label{tab:sim_data_dev}
\end{table}

\begin{table}
	\begin{center}
	\begin{tabular}{l| rrrr | rrrr | rrrr}		\hline \hline
		& \multicolumn{4}{c|}{IID errors} & \multicolumn{4}{c|}{MA errors} & \multicolumn{4}{c}{AR errors} \\
		$n$ & $\mu_1$ & $\mu_2$ & $\mu_3$ & $\mu_4$ & $\mu_1$ & $\mu_2$ & $\mu_3$ & $\mu_4$ & $\mu_1$ & $\mu_2$ & $\mu_3$ & $\mu_4$ \\ 
		\hline 
		\addlinespace[.2cm]
		\multicolumn{13}{l}{\quad\textit{Panel A: empirical rejection rates of mon. scheme \eqref{eq:cusum_test}}} \\ 
		40 & 3.5 & 100.0 & 100.0 & 100.0 & 12.0 & 99.9 & 100.0 & 100.0 & 10.6 & 99.8 & 100.0 & 100.0\\
		100 & 2.9 & 100.0 & 100.0 & 100.0 & 11.1 & 100.0 & 100.0 & 100.0 & 9.5 & 100.0 & 100.0 & 100.0\\
		200 & 3.4 & 100.0 & 100.0 & 100.0 & 12.9 & 100.0 & 100.0 & 100.0 & 11.5 & 100.0 & 100.0 & 100.0\\
		\addlinespace[.2cm]
		\multicolumn{13}{l}{\quad\textit{Panel B: empirical rejection rates of mon. scheme \eqref{eq:page_cusum_test}}} \\ 
		40 & 3.5 & 100.0 & 100.0 & 100.0 & 12.5 & 100.0 & 100.0 & 100.0 & 10.9 & 100.0 & 100.0 & 100.0\\
		100 & 3.0 & 100.0 & 100.0 & 100.0 & 13.1 & 100.0 & 100.0 & 100.0 & 10.4 & 100.0 & 100.0 & 100.0\\
		200 & 3.6 & 100.0 & 100.0 & 100.0 & 13.5 & 100.0 & 100.0 & 100.0 & 12.8 & 100.0 & 100.0 & 100.0\\
		\addlinespace[.2cm]
		\multicolumn{13}{l}{\quad\textit{Panel C: average time of first deviation of mon. scheme \eqref{eq:cusum_test}}} \\ 
		40 & 2.79 & 3.19 & 2.45 & 1.08 & 2.70 & 3.10 & 2.43 & 1.08 & 2.79 & 3.14 & 2.43 & 1.08\\
		100 & 2.97 & 2.52 & 2.25 & 1.05 & 2.59 & 2.48 & 2.22 & 1.05 & 2.95 & 2.49 & 2.23 & 1.05\\
		200 & 3.17 & 2.16 & 2.11 & 1.04 & 2.78 & 2.07 & 2.09 & 1.04 & 2.64 & 2.09 & 2.10 & 1.04\\
		\addlinespace[.2cm]
		\multicolumn{13}{l}{\quad\textit{Panel D: average time of first deviation of mon. scheme \eqref{eq:page_cusum_test}}} \\ 
		40 & 2.88 & 3.08 & 2.40 & 1.11 & 2.79 & 2.96 & 2.36 & 1.11 & 2.77 & 3.00 & 2.36 & 1.11\\
		100 & 3.01 & 2.34 & 2.21 & 1.06 & 2.61 & 2.28 & 2.16 & 1.06 & 2.95 & 2.30 & 2.17 & 1.06\\
		200 & 3.21 & 2.09 & 2.07 & 1.04 & 2.77 & 1.97 & 2.04 & 1.04 & 2.64 & 2.00 & 2.05 & 1.04\\
		\hline \hline
	\end{tabular} \medskip
\end{center}
	\caption{Empirical rejection rates and average times of the first deviation of CUSUM-based monitoring schemes for the hypotheses \eqref{eq:null}, different mean functions $\mu$, different error processes, and sample sizes $n=40, 100, 200$.}
	\label{tab:sim_data_cusum}
\end{table}

\begin{table}
	\footnotesize
	\begin{center}
	\begin{tabular}{l| rrr | rrr | rrr | rrr | rrr }	\hline \hline
	&\multicolumn{3}{c|}{mon. scheme \eqref{eq:naive_scheme} }&\multicolumn{3}{c|}{mon. scheme \eqref{eq:t_test}} & \multicolumn{3}{c|}{mon. scheme \eqref{eq:t_test_corr}}&\multicolumn{3}{c|}{mon. scheme \eqref{eq:mon_scheme}}&\multicolumn{3}{c}{mon. scheme \eqref{eq:approx_mon_scheme}}\\
	$\Delta$ & 40 & 100 & 200 & 40 & 100 & 200 & 40 & 100 & 200 & 40 & 100 & 200 & 40 & 100 & 200 \\ 
	\hline 
	\addlinespace[.2cm]
	\multicolumn{16}{l}{\quad\textit{Panel A: $\mu_1$}} \\ 
	\textbf{0} & \textbf{100.0} & \textbf{100.0} & \textbf{100.0} & \textbf{96.1} & \textbf{96.8} & \textbf{96.0} & \textbf{44.8} & \textbf{37.9} & \textbf{33.4} & \textbf{20.6} & \textbf{13.1} & \textbf{6.8} & \textbf{58.8} & \textbf{50.5} & \textbf{38.9}\\
	1 & 100.0 & 100.0 & 100.0 & 65.7 & 39.6 & 17.4 & 14.5 & 4.1 & 0.7 & 18.1 & 7.2 & 1.3 & 24.6 & 9.4 & 2.0\\
	2 & 100.0 & 100.0 & 100.0 & 20.5 & 4.0 & 0.2 & 2.9 & 0.1 & 0.0 & 7.8 & 1.5 & 0.1 & 10.7 & 1.8 & 0.1\\
	10 & 99.2 & 100.0 & 100.0 & 0.0 & 0.0 & 0.0 & 0.0 & 0.0 & 0.0 & 0.0 & 0.0 & 0.0 & 0.0 & 0.0 & 0.0\\
	15 & 22.8 & 40.2 & 64.6 & 0.0 & 0.0 & 0.0 & 0.0 & 0.0 & 0.0 & 0.0 & 0.0 & 0.0 & 0.0 & 0.0 & 0.0\\
	20 & 0.8 & 1.1 & 2.1 & 0.0 & 0.0 & 0.0 & 0.0 & 0.0 & 0.0 & 0.0 & 0.0 & 0.0 & 0.0 & 0.0 & 0.0\\

	\addlinespace[.2cm]
	\multicolumn{16}{l}{\quad\textit{Panel B: $\mu_2$}} \\ 
	12 & 100.0 & 100.0 & 100.0 & 100.0 & 100.0 & 100.0 & 100.0 & 100.0 & 100.0 & 99.3 & 100.0 & 100.0 & 99.7 & 100.0 & 100.0\\
	14 & 100.0 & 100.0 & 100.0 & 100.0 & 100.0 & 100.0 & 99.9 & 100.0 & 100.0 & 94.8 & 100.0 & 100.0 & 97.6 & 100.0 & 100.0\\
	16 & 100.0 & 100.0 & 100.0 & 99.4 & 100.0 & 100.0 & 95.3 & 99.9 & 100.0 & 72.9 & 91.7 & 100.0 & 81.5 & 96.6 & 100.0\\
	18 & 100.0 & 100.0 & 100.0 & 86.0 & 97.2 & 99.9 & 54.6 & 77.6 & 94.7 & 34.2 & 49.2 & 63.6 & 43.4 & 59.7 & 78.1\\
	19 & 100.0 & 100.0 & 100.0 & 65.4 & 77.3 & 88.0 & 26.7 & 37.6 & 47.1 & 18.4 & 22.6 & 22.3 & 23.4 & 30.4 & 30.2\\
	\textbf{20} & \textbf{100.0} & \textbf{100.0} & \textbf{100.0} & \textbf{35.5} & \textbf{36.9} & \textbf{33.0} & \textbf{9.5} & \textbf{7.3} & \textbf{5.7} & \textbf{9.1} & \textbf{5.9} & \textbf{2.7} & \textbf{11.6} & \textbf{9.2} & \textbf{5.8}\\
	21 & 100.0 & 100.0 & 100.0 & 11.8 & 7.2 & 2.5 & 1.7 & 0.5 & 0.3 & 3.6 & 1.4 & 0.0 & 5.1 & 1.9 & 0.0\\

	\addlinespace[.2cm]
	\multicolumn{16}{l}{\quad\textit{Panel C: $\mu_3$}} \\ 
	6 & 100.0 & 100.0 & 100.0 & 98.5 & 100.0 & 100.0 & 81.4 & 99.4 & 100.0 & 93.9 & 100.0 & 100.0 & 96.0 & 100.0 & 100.0\\
	8 & 100.0 & 100.0 & 100.0 & 77.6 & 95.9 & 100.0 & 30.1 & 65.3 & 94.7 & 77.1 & 99.0 & 100.0 & 82.5 & 99.3 & 100.0\\
	10 & 100.0 & 100.0 & 100.0 & 25.1 & 37.9 & 49.2 & 3.9 & 5.0 & 6.6 & 48.4 & 84.6 & 98.8 & 54.2 & 87.4 & 99.4\\
	12 & 100.0 & 100.0 & 100.0 & 2.4 & 1.1 & 0.3 & 0.2 & 0.0 & 0.0 & 21.9 & 44.3 & 69.2 & 25.7 & 47.6 & 72.8\\
	14 & 100.0 & 100.0 & 100.0 & 0.1 & 0.0 & 0.0 & 0.0 & 0.0 & 0.0 & 5.6 & 9.4 & 11.4 & 6.8 & 10.6 & 12.9\\
	15 & 100.0 & 100.0 & 100.0 & 0.1 & 0.0 & 0.0 & 0.0 & 0.0 & 0.0 & 2.7 & 2.8 & 2.4 & 2.8 & 3.3 & 2.9\\
	\textbf{16} & \textbf{100.0} & \textbf{100.0} & \textbf{100.0} & \textbf{0.0} & \textbf{0.0} & \textbf{0.0} & \textbf{0.0} & \textbf{0.0} & \textbf{0.0} & \textbf{1.3} & \textbf{0.7} & \textbf{0.2} & \textbf{1.3} & \textbf{0.9} & \textbf{0.2}\\

	\addlinespace[.2cm]
	\multicolumn{16}{l}{\quad\textit{Panel D: $\mu_4$}} \\ 
	14 & 100.0 & 100.0 & 100.0 & 100.0 & 100.0 & 100.0 & 100.0 & 100.0 & 100.0 & 98.6 & 100.0 & 100.0 & 99.7 & 100.0 & 100.0\\
	16 & 100.0 & 100.0 & 100.0 & 100.0 & 100.0 & 100.0 & 98.9 & 100.0 & 100.0 & 88.3 & 99.5 & 100.0 & 93.9 & 99.7 & 100.0\\
	18 & 100.0 & 100.0 & 100.0 & 97.2 & 99.6 & 100.0 & 80.9 & 93.8 & 99.5 & 59.8 & 74.9 & 91.6 & 68.7 & 86.2 & 96.8\\
	19 & 100.0 & 100.0 & 100.0 & 87.0 & 93.2 & 98.7 & 54.3 & 64.2 & 76.7 & 40.6 & 42.4 & 50.0 & 49.2 & 57.8 & 64.5\\
	\textbf{20} & \textbf{100.0} & \textbf{100.0} & \textbf{100.0} & \textbf{62.6} & \textbf{62.6} & \textbf{62.1} & \textbf{25.4} & \textbf{16.9} & \textbf{13.8} & \textbf{23.4} & \textbf{17.6} & \textbf{10.2} & \textbf{29.1} & \textbf{25.4} & \textbf{18.9}\\
	21 & 100.0 & 100.0 & 100.0 & 33.4 & 15.8 & 6.1 & 7.2 & 1.2 & 0.2 & 10.4 & 3.3 & 0.6 & 14.0 & 5.1 & 1.8\\
	22 & 100.0 & 100.0 & 100.0 & 10.3 & 0.8 & 0.1 & 0.8 & 0.0 & 0.0 & 4.5 & 0.7 & 0.0 & 5.4 & 0.8 & 0.0\\

	\hline \hline
\end{tabular}
	\end{center}
	\medskip
	\caption{Empirical rejection rates of various monitoring schemes for the hypotheses  \eqref{eq:null}, different mean functions $\mu$, MA errors and sample sizes $n=40, 100, 200$. }
	\label{tab:sim_data_ma}
\end{table}

\begin{table}
	\footnotesize
	\begin{center}
	\begin{tabular}{l| rrr | rrr | rrr | rrr | rrr }	\hline \hline
	&\multicolumn{3}{c|}{mon. scheme \eqref{eq:naive_scheme} }&\multicolumn{3}{c|}{mon. scheme \eqref{eq:t_test}} & \multicolumn{3}{c|}{mon. scheme \eqref{eq:t_test_corr}}&\multicolumn{3}{c|}{mon. scheme \eqref{eq:mon_scheme}}&\multicolumn{3}{c}{mon. scheme \eqref{eq:approx_mon_scheme}}\\
	$\Delta$ & 40 & 100 & 200 & 40 & 100 & 200 & 40 & 100 & 200 & 40 & 100 & 200 & 40 & 100 & 200 \\ 
	\hline 
	\addlinespace[.2cm]
	\multicolumn{16}{l}{\quad\textit{Panel A: $\mu_1$}} \\ 
	\textbf{0} & \textbf{100.0} & \textbf{100.0} & \textbf{100.0} & \textbf{95.2} & \textbf{95.8} & \textbf{94.5} & \textbf{39.9} & \textbf{32.3} & \textbf{26.6} & \textbf{20.4} & \textbf{17.6} & \textbf{9.5} & \textbf{62.0} & \textbf{61.0} & \textbf{46.7}\\
	1 & 100.0 & 100.0 & 100.0 & 60.0 & 33.9 & 14.1 & 11.8 & 3.1 & 1.1 & 16.9 & 9.1 & 2.0 & 23.6 & 13.1 & 3.2\\
	2 & 100.0 & 100.0 & 100.0 & 17.1 & 2.7 & 0.4 & 1.9 & 0.0 & 0.0 & 5.5 & 0.8 & 0.0 & 8.1 & 1.2 & 0.0\\
	10 & 99.3 & 100.0 & 100.0 & 0.0 & 0.0 & 0.0 & 0.0 & 0.0 & 0.0 & 0.0 & 0.0 & 0.0 & 0.0 & 0.0 & 0.0\\
	15 & 20.8 & 41.4 & 66.1 & 0.0 & 0.0 & 0.0 & 0.0 & 0.0 & 0.0 & 0.0 & 0.0 & 0.0 & 0.0 & 0.0 & 0.0\\
	20 & 0.7 & 0.8 & 3.0 & 0.0 & 0.0 & 0.0 & 0.0 & 0.0 & 0.0 & 0.0 & 0.0 & 0.0 & 0.0 & 0.0 & 0.0\\

	\addlinespace[.2cm]
	\multicolumn{16}{l}{\quad\textit{Panel B: $\mu_2$}} \\ 
	12 & 100.0 & 100.0 & 100.0 & 100.0 & 100.0 & 100.0 & 100.0 & 100.0 & 100.0 & 99.7 & 100.0 & 100.0 & 100.0 & 100.0 & 100.0\\
	14 & 100.0 & 100.0 & 100.0 & 100.0 & 100.0 & 100.0 & 100.0 & 100.0 & 100.0 & 95.0 & 100.0 & 100.0 & 97.9 & 100.0 & 100.0\\
	16 & 100.0 & 100.0 & 100.0 & 99.5 & 100.0 & 100.0 & 94.3 & 100.0 & 100.0 & 77.9 & 97.4 & 100.0 & 86.3 & 99.3 & 100.0\\
	18 & 100.0 & 100.0 & 100.0 & 87.8 & 97.6 & 100.0 & 57.2 & 79.0 & 95.0 & 38.1 & 58.9 & 75.3 & 48.2 & 69.9 & 84.4\\
	19 & 100.0 & 100.0 & 100.0 & 65.6 & 78.9 & 88.8 & 28.9 & 33.4 & 48.3 & 21.5 & 27.7 & 28.5 & 27.2 & 36.3 & 41.0\\
	\textbf{20} & \textbf{100.0} & \textbf{100.0} & \textbf{100.0} & \textbf{37.3} & \textbf{33.0} & \textbf{34.5} & \textbf{9.0} & \textbf{7.0} & \textbf{4.3} & \textbf{10.5} & \textbf{7.6} & \textbf{4.1} & \textbf{13.4} & \textbf{12.1} & \textbf{6.4}\\
	21 & 100.0 & 100.0 & 100.0 & 13.3 & 6.9 & 1.6 & 2.7 & 0.4 & 0.1 & 5.1 & 1.6 & 0.0 & 6.0 & 1.9 & 0.0\\

	\addlinespace[.2cm]
	\multicolumn{16}{l}{\quad\textit{Panel C: $\mu_3$}} \\ 
	6 & 100.0 & 100.0 & 100.0 & 98.0 & 100.0 & 100.0 & 79.5 & 99.7 & 100.0 & 96.6 & 100.0 & 100.0 & 98.0 & 100.0 & 100.0\\
	8 & 100.0 & 100.0 & 100.0 & 76.6 & 97.4 & 99.9 & 30.0 & 65.4 & 92.6 & 82.7 & 99.5 & 100.0 & 87.9 & 99.7 & 100.0\\
	10 & 100.0 & 100.0 & 100.0 & 25.6 & 36.8 & 51.6 & 2.5 & 2.7 & 5.8 & 54.5 & 89.0 & 99.8 & 59.7 & 90.7 & 99.8\\
	12 & 100.0 & 100.0 & 100.0 & 2.0 & 0.6 & 0.2 & 0.0 & 0.0 & 0.0 & 26.0 & 52.7 & 79.5 & 28.9 & 55.2 & 81.9\\
	14 & 100.0 & 100.0 & 100.0 & 0.0 & 0.0 & 0.0 & 0.0 & 0.0 & 0.0 & 8.3 & 11.6 & 16.5 & 9.3 & 13.5 & 18.6\\
	15 & 100.0 & 100.0 & 100.0 & 0.0 & 0.0 & 0.0 & 0.0 & 0.0 & 0.0 & 3.8 & 4.2 & 4.0 & 4.4 & 4.8 & 4.9\\
	\textbf{16} & \textbf{100.0} & \textbf{100.0} & \textbf{100.0} & \textbf{0.0} & \textbf{0.0} & \textbf{0.0} & \textbf{0.0} & \textbf{0.0} & \textbf{0.0} & \textbf{1.6} & \textbf{1.0} & \textbf{0.5} & \textbf{1.7} & \textbf{1.1} & \textbf{0.5}\\

	\addlinespace[.2cm]
	\multicolumn{16}{l}{\quad\textit{Panel D: $\mu_4$}} \\ 
	14 & 100.0 & 100.0 & 100.0 & 100.0 & 100.0 & 100.0 & 100.0 & 100.0 & 100.0 & 99.3 & 100.0 & 100.0 & 99.8 & 100.0 & 100.0\\
	16 & 100.0 & 100.0 & 100.0 & 100.0 & 100.0 & 100.0 & 99.9 & 100.0 & 100.0 & 93.2 & 99.7 & 100.0 & 96.1 & 100.0 & 100.0\\
	18 & 100.0 & 100.0 & 100.0 & 97.7 & 99.5 & 100.0 & 79.6 & 94.3 & 99.2 & 65.5 & 84.1 & 94.5 & 74.6 & 91.5 & 98.2\\
	19 & 100.0 & 100.0 & 100.0 & 86.9 & 94.2 & 98.5 & 52.0 & 60.2 & 74.8 & 44.1 & 53.0 & 58.6 & 53.0 & 67.6 & 75.0\\
	\textbf{20} & \textbf{100.0} & \textbf{100.0} & \textbf{100.0} & \textbf{62.2} & \textbf{59.0} & \textbf{60.4} & \textbf{22.7} & \textbf{14.1} & \textbf{13.4} & \textbf{25.5} & \textbf{21.2} & \textbf{14.8} & \textbf{31.1} & \textbf{27.8} & \textbf{22.6}\\
	21 & 100.0 & 100.0 & 100.0 & 30.4 & 13.8 & 6.4 & 5.3 & 1.2 & 0.3 & 13.2 & 4.5 & 1.3 & 16.5 & 7.3 & 2.3\\
	22 & 100.0 & 100.0 & 100.0 & 9.5 & 1.0 & 0.0 & 1.4 & 0.0 & 0.0 & 6.3 & 0.7 & 0.0 & 7.1 & 1.0 & 0.1\\
	\hline \hline
\end{tabular}
	\end{center}
	\medskip
	\caption{Empirical rejection rates of various monitoring schemes for the hypotheses  \eqref{eq:null}, different mean functions $\mu$, AR errors and sample sizes $n=40, 100, 200$. }
	\label{tab:sim_data_ar}
\end{table}

\clearpage
\pagebreak

\subsection{Additional Results of the Real Data Experiments}
\label{sec:app_real_data}

\begin{figure}[h]
	\vskip -0.1in
	\begin{center}
		\includegraphics[width=0.45\textwidth]{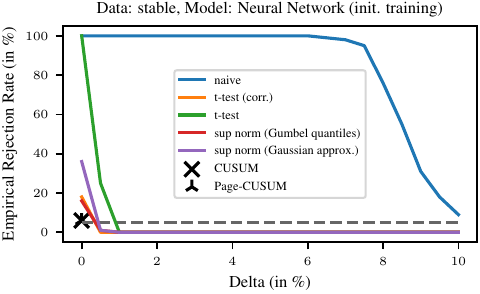} \hspace{0.1cm}
		\includegraphics[width=0.45\textwidth]{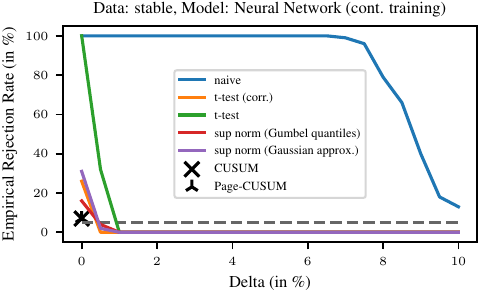} \\
		\includegraphics[width=0.45\textwidth]{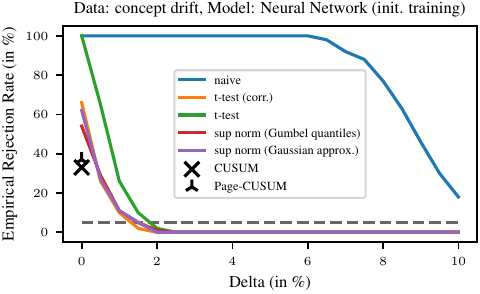} \hspace{0.1cm}
		\includegraphics[width=0.45\textwidth]{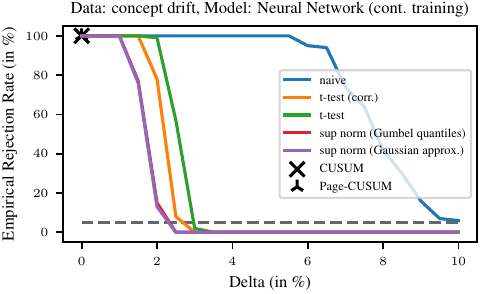} \\
		\includegraphics[width=0.45\textwidth]{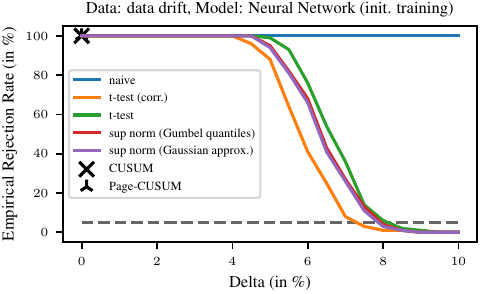} \hspace{0.1cm}
		\includegraphics[width=0.45\textwidth]{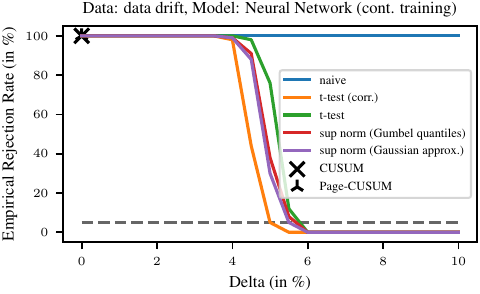} \\
		\vskip -0.1in
		\caption{Empirical rejection rates of the different monitoring schemes for neural networks with initial and continuous training and varying values of $\Delta$ based on the models' \textit{accuracies}. The dashed horizontal line marks the level $\alpha=5\%$. Top left: \NNI~ with stable data. Top right: \NNC~ with stable data. Center left: \NNI~ and data with concept drift. Center right: \NNC~ and data with concept drift. Bottom left: \NNI~ and data with data drift. Bottom right: \NNC~ and data with data drift.}
		\label{fig:real_data_nn}
	\end{center}
\end{figure}

\begin{figure}
	\vskip -0.1in
	\begin{center}
		\includegraphics[width=0.45\textwidth]{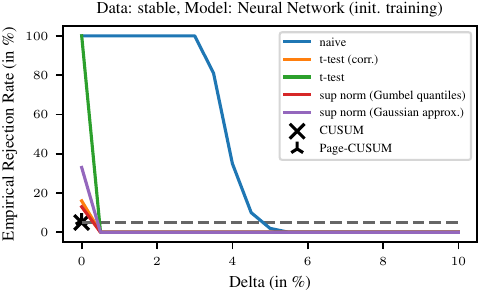} \hspace{0.1cm}
		\includegraphics[width=0.45\textwidth]{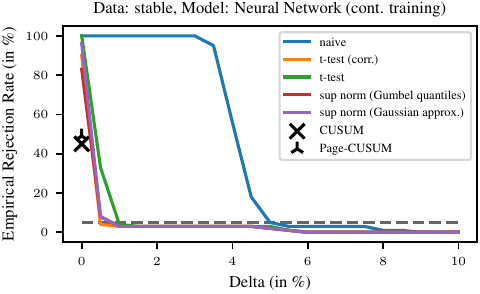}\\
		\includegraphics[width=0.45\textwidth]{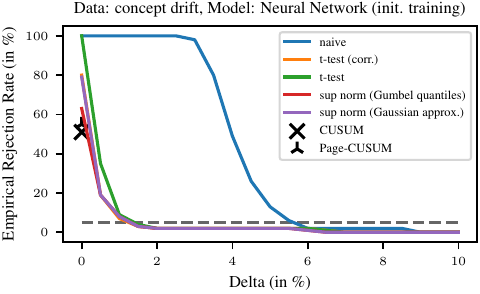} \hspace{0.1cm}
		\includegraphics[width=0.45\textwidth]{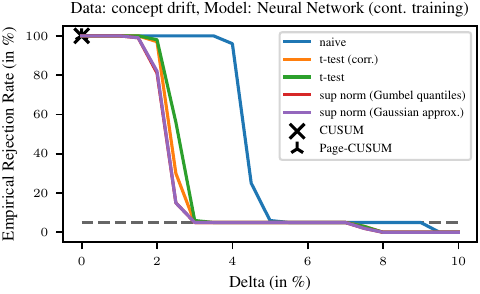} \\
		\includegraphics[width=0.45\textwidth]{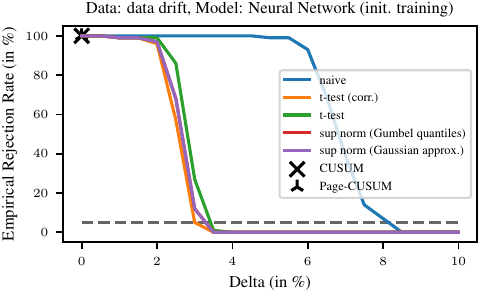} \hspace{0.1cm}
		\includegraphics[width=0.45\textwidth]{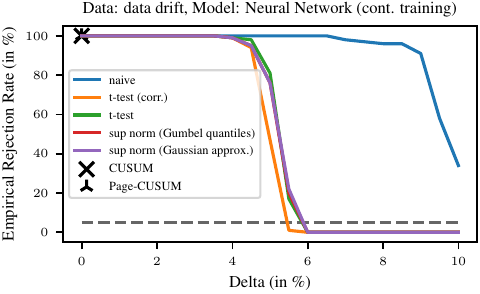} \\
		\vskip -0.1in
		\caption{Empirical rejection rates of the different monitoring schemes for neural networks with initial and continuous training and varying values of $\Delta$ based on the models' \textit{confidences}. The dashed horizontal line marks the level $\alpha=5\%$. Top left: \NNI~ with stable data. Top right: \NNC~ with stable data. Center left: \NNI~ and data with concept drift. Center right: \NNC~ and data with concept drift. Bottom left: \NNI~ and data with data drift. Bottom right: \NNC~ and data with data drift.}
		\label{fig:real_data_nn_conf}
	\end{center}
\end{figure}

\begin{table}[h!]
	\scriptsize
	\begin{center}
	\begin{tabular}{r| rrrrrrrrrrr | rrrrrrrrrrr }	\hline \hline
	&\multicolumn{11}{c|}{Accuracy}&\multicolumn{11}{c}{Confidence}\\
	$\Delta$ & 0 & 1 & 2 & 3 & 4 & 5 & 6 & 7 & 8 & 9 & 10 & 0 & 1 & 2 & 3 & 4 & 5 & 6 & 7 & 8 & 9 & 10 \\ 
	\hline 
	\addlinespace[.1cm]
	\multicolumn{23}{l}{\quad\textit{Panel A: Logistic Regression, stable data}} \\ 
	\eqref{eq:naive_scheme} & 100 & 100 & 100 & 100 & 100 & 100 & 100 & 98 & 75 & 36 & 11 & 100 & 100 & 100 & 100 & 43 & 4 & 0 & 0 & 0 & 0 & 0\\
	\eqref{eq:t_test} & 100 & 0 & 0 & 0 & 0 & 0 & 0 & 0 & 0 & 0 & 0 & 100 & 0 & 0 & 0 & 0 & 0 & 0 & 0 & 0 & 0 & 0\\
	\eqref{eq:t_test_corr}& 12 & 0 & 0 & 0 & 0 & 0 & 0 & 0 & 0 & 0 & 0 & 20 & 0 & 0 & 0 & 0 & 0 & 0 & 0 & 0 & 0 & 0\\
	\eqref{eq:mon_scheme} & 8 & 0 & 0 & 0 & 0 & 0 & 0 & 0 & 0 & 0 & 0 & 10 & 0 & 0 & 0 & 0 & 0 & 0 & 0 & 0 & 0 & 0\\
	\eqref{eq:approx_mon_scheme} & 19 & 0 & 0 & 0 & 0 & 0 & 0 & 0 & 0 & 0 & 0 & 30 & 0 & 0 & 0 & 0 & 0 & 0 & 0 & 0 & 0 & 0\\
	
	\addlinespace[.1cm]
	\multicolumn{23}{l}{\quad\textit{Panel B: Neural Network (initial training), stable data}} \\ 
	\eqref{eq:naive_scheme} & 100 & 100 & 100 & 100 & 100 & 100 & 100 & 98 & 76 & 31 & 9 & 100 & 100 & 100 & 100 & 35 & 2 & 0 & 0 & 0 & 0 & 0\\
	\eqref{eq:t_test} & 100 & 0 & 0 & 0 & 0 & 0 & 0 & 0 & 0 & 0 & 0 & 100 & 0 & 0 & 0 & 0 & 0 & 0 & 0 & 0 & 0 & 0\\
	\eqref{eq:t_test_corr}& 18 & 0 & 0 & 0 & 0 & 0 & 0 & 0 & 0 & 0 & 0 & 16 & 0 & 0 & 0 & 0 & 0 & 0 & 0 & 0 & 0 & 0\\
	\eqref{eq:mon_scheme} & 16 & 0 & 0 & 0 & 0 & 0 & 0 & 0 & 0 & 0 & 0 & 13 & 0 & 0 & 0 & 0 & 0 & 0 & 0 & 0 & 0 & 0\\
	\eqref{eq:approx_mon_scheme} & 36 & 0 & 0 & 0 & 0 & 0 & 0 & 0 & 0 & 0 & 0 & 33 & 0 & 0 & 0 & 0 & 0 & 0 & 0 & 0 & 0 & 0\\
	
	\addlinespace[.1cm]
	\multicolumn{23}{l}{\quad\textit{Panel C: Neural Network (continuous training), stable data}} \\ 
	\eqref{eq:naive_scheme} & 100 & 100 & 100 & 100 & 100 & 100 & 100 & 99 & 79 & 40 & 13 & 100 & 100 & 100 & 100 & 56 & 5 & 3 & 3 & 1 & 0 & 0\\
	\eqref{eq:t_test} & 100 & 0 & 0 & 0 & 0 & 0 & 0 & 0 & 0 & 0 & 0 & 100 & 4 & 3 & 3 & 3 & 3 & 0 & 0 & 0 & 0 & 0\\
	\eqref{eq:t_test_corr}& 26 & 0 & 0 & 0 & 0 & 0 & 0 & 0 & 0 & 0 & 0 & 90 & 3 & 3 & 3 & 3 & 2 & 0 & 0 & 0 & 0 & 0\\
	\eqref{eq:mon_scheme} & 16 & 0 & 0 & 0 & 0 & 0 & 0 & 0 & 0 & 0 & 0 & 83 & 3 & 3 & 3 & 3 & 2 & 0 & 0 & 0 & 0 & 0\\
	\eqref{eq:approx_mon_scheme} & 31 & 0 & 0 & 0 & 0 & 0 & 0 & 0 & 0 & 0 & 0 & 96 & 3 & 3 & 3 & 3 & 2 & 0 & 0 & 0 & 0 & 0\\

	\addlinespace[.1cm]
	\multicolumn{23}{l}{\quad\textit{Panel D: Logistic Regression, data with concept drift}} \\ 
	\eqref{eq:naive_scheme} & 100 & 100 & 100 & 100 & 100 & 100 & 100 & 95 & 69 & 42 & 21 & 100 & 100 & 100 & 100 & 42 & 0 & 0 & 0 & 0 & 0 & 0\\
	\eqref{eq:t_test} & 100 & 10 & 0 & 0 & 0 & 0 & 0 & 0 & 0 & 0 & 0 & 100 & 0 & 0 & 0 & 0 & 0 & 0 & 0 & 0 & 0 & 0\\
	\eqref{eq:t_test_corr}& 45 & 0 & 0 & 0 & 0 & 0 & 0 & 0 & 0 & 0 & 0 & 51 & 0 & 0 & 0 & 0 & 0 & 0 & 0 & 0 & 0 & 0\\
	\eqref{eq:mon_scheme} & 35 & 0 & 0 & 0 & 0 & 0 & 0 & 0 & 0 & 0 & 0 & 37 & 0 & 0 & 0 & 0 & 0 & 0 & 0 & 0 & 0 & 0\\
	\eqref{eq:approx_mon_scheme} & 49 & 0 & 0 & 0 & 0 & 0 & 0 & 0 & 0 & 0 & 0 & 63 & 0 & 0 & 0 & 0 & 0 & 0 & 0 & 0 & 0 & 0\\
	
	\addlinespace[.1cm]
	\multicolumn{23}{l}{\quad\textit{Panel E: Neural Network (initial training), data with concept drift}} \\ 
	\eqref{eq:naive_scheme} & 100 & 100 & 100 & 100 & 100 & 100 & 100 & 92 & 77 & 46 & 18 & 100 & 100 & 100 & 98 & 49 & 13 & 2 & 2 & 2 & 0 & 0\\
	\eqref{eq:t_test} & 100 & 26 & 2 & 0 & 0 & 0 & 0 & 0 & 0 & 0 & 0 & 100 & 9 & 2 & 2 & 2 & 2 & 2 & 0 & 0 & 0 & 0\\
	\eqref{eq:t_test_corr}& 66 & 10 & 0 & 0 & 0 & 0 & 0 & 0 & 0 & 0 & 0 & 80 & 7 & 2 & 2 & 2 & 2 & 2 & 0 & 0 & 0 & 0\\
	\eqref{eq:mon_scheme} & 54 & 11 & 1 & 0 & 0 & 0 & 0 & 0 & 0 & 0 & 0 & 63 & 8 & 2 & 2 & 2 & 2 & 1 & 0 & 0 & 0 & 0\\
	\eqref{eq:approx_mon_scheme} & 62 & 11 & 0 & 0 & 0 & 0 & 0 & 0 & 0 & 0 & 0 & 79 & 8 & 2 & 2 & 2 & 2 & 1 & 0 & 0 & 0 & 0\\
	
	\addlinespace[.1cm]
	\multicolumn{23}{l}{\quad\textit{Panel F: Neural Network (continuous training), data with concept drift}} \\ 
	\eqref{eq:naive_scheme} & 100 & 100 & 100 & 100 & 100 & 100 & 95 & 74 & 42 & 16 & 6 & 100 & 100 & 100 & 100 & 96 & 6 & 5 & 5 & 5 & 5 & 0\\
	\eqref{eq:t_test} & 100 & 100 & 99 & 2 & 0 & 0 & 0 & 0 & 0 & 0 & 0 & 100 & 100 & 98 & 6 & 5 & 5 & 5 & 5 & 0 & 0 & 0\\
	\eqref{eq:t_test_corr}& 100 & 100 & 78 & 0 & 0 & 0 & 0 & 0 & 0 & 0 & 0 & 100 & 100 & 97 & 5 & 5 & 5 & 5 & 5 & 0 & 0 & 0\\
	\eqref{eq:mon_scheme} & 100 & 100 & 15 & 0 & 0 & 0 & 0 & 0 & 0 & 0 & 0 & 100 & 100 & 81 & 5 & 5 & 5 & 5 & 5 & 0 & 0 & 0\\
	\eqref{eq:approx_mon_scheme} & 100 & 100 & 13 & 0 & 0 & 0 & 0 & 0 & 0 & 0 & 0 & 100 & 100 & 82 & 5 & 5 & 5 & 5 & 5 & 0 & 0 & 0\\

	\addlinespace[.1cm]
	\multicolumn{23}{l}{\quad\textit{Panel G: Logistic Regression, data with data drift}} \\ 
	\eqref{eq:naive_scheme} & 100 & 100 & 100 & 100 & 100 & 100 & 100 & 100 & 100 & 100 & 100 & 100 & 100 & 100 & 100 & 100 & 100 & 100 & 67 & 10 & 2 & 1\\
	\eqref{eq:t_test} & 100 & 100 & 100 & 100 & 100 & 100 & 94 & 25 & 0 & 0 & 0 & 100 & 100 & 100 & 57 & 0 & 0 & 0 & 0 & 0 & 0 & 0\\
	\eqref{eq:t_test_corr}& 100 & 100 & 100 & 100 & 100 & 98 & 44 & 0 & 0 & 0 & 0 & 100 & 100 & 100 & 7 & 0 & 0 & 0 & 0 & 0 & 0 & 0\\
	\eqref{eq:mon_scheme} & 100 & 100 & 100 & 100 & 100 & 100 & 83 & 17 & 1 & 0 & 0 & 100 & 100 & 100 & 15 & 0 & 0 & 0 & 0 & 0 & 0 & 0\\
	\eqref{eq:approx_mon_scheme} & 100 & 100 & 100 & 100 & 100 & 99 & 80 & 16 & 1 & 0 & 0 & 100 & 100 & 100 & 15 & 0 & 0 & 0 & 0 & 0 & 0 & 0\\

	\addlinespace[.1cm]
	\multicolumn{23}{l}{\quad\textit{Panel H: Neural Network (initial training), data with data drift}} \\ 
	\eqref{eq:naive_scheme} & 100 & 100 & 100 & 100 & 100 & 100 & 100 & 100 & 100 & 100 & 100 & 100 & 100 & 100 & 100 & 100 & 99 & 93 & 39 & 7 & 0 & 0\\
	\eqref{eq:t_test} & 100 & 100 & 100 & 100 & 100 & 99 & 76 & 36 & 6 & 1 & 0 & 100 & 99 & 99 & 27 & 0 & 0 & 0 & 0 & 0 & 0 & 0\\
	\eqref{eq:t_test_corr}& 100 & 100 & 100 & 100 & 100 & 88 & 41 & 8 & 1 & 0 & 0 & 100 & 99 & 96 & 5 & 0 & 0 & 0 & 0 & 0 & 0 & 0\\
	\eqref{eq:mon_scheme} & 100 & 100 & 100 & 100 & 100 & 95 & 68 & 27 & 4 & 0 & 0 & 100 & 99 & 97 & 12 & 0 & 0 & 0 & 0 & 0 & 0 & 0\\
	\eqref{eq:approx_mon_scheme} & 100 & 100 & 100 & 100 & 100 & 94 & 66 & 26 & 3 & 0 & 0 & 100 & 99 & 97 & 12 & 0 & 0 & 0 & 0 & 0 & 0 & 0\\
	
	\addlinespace[.1cm]
	\multicolumn{23}{l}{\quad\textit{Panel I: Neural Network (continuous training), data with data drift}} \\ 
	\eqref{eq:naive_scheme} & 100 & 100 & 100 & 100 & 100 & 100 & 100 & 100 & 100 & 100 & 100 & 100 & 100 & 100 & 100 & 100 & 100 & 100 & 98 & 96 & 91 & 34\\
	\eqref{eq:t_test} & 100 & 100 & 100 & 100 & 100 & 76 & 0 & 0 & 0 & 0 & 0 & 100 & 100 & 100 & 100 & 99 & 81 & 0 & 0 & 0 & 0 & 0\\
	\eqref{eq:t_test_corr}& 100 & 100 & 100 & 100 & 98 & 5 & 0 & 0 & 0 & 0 & 0 & 100 & 100 & 100 & 100 & 99 & 47 & 0 & 0 & 0 & 0 & 0\\
	\eqref{eq:mon_scheme} & 100 & 100 & 100 & 100 & 99 & 38 & 0 & 0 & 0 & 0 & 0 & 100 & 100 & 100 & 100 & 99 & 76 & 0 & 0 & 0 & 0 & 0\\
	\eqref{eq:approx_mon_scheme} & 100 & 100 & 100 & 100 & 99 & 30 & 0 & 0 & 0 & 0 & 0 & 100 & 100 & 100 & 100 & 99 & 76 & 0 & 0 & 0 & 0 & 0\\
	
	\hline \hline
\end{tabular}
	\end{center}
	\medskip
	\caption{Empirical rejection rates of various monitoring schemes based on the models' accuracies and confidences for the hypotheses \eqref{eq:null}, the models \LogReg, \NNI~ and \NNC~ and different data regimes.}
	\label{tab:real_data}
\end{table}

\begin{table}[h!]
	\small
	\begin{center}
	\begin{tabular}{l| rrr | rrr | rrr}		\hline \hline
		& \multicolumn{3}{c|}{Stable} & \multicolumn{3}{c|}{Concept drift} & \multicolumn{3}{c}{Data drift} \\
		& LogReg & NN (init) & NN (cont) & LogReg & NN (init) & NN (cont) & LogReg & NN (init) & NN (cont) \\ 
		\hline 
		\addlinespace[.2cm]
		\multicolumn{10}{l}{\quad\textit{Panel A: empirical rejection rates of mon. scheme \eqref{eq:cusum_test}}} \\ 
		acc. & 5 & 5 & 4 & 10 & 8 & 68 & 100 & 100 & 100\\
		conf. & 7 & 5 & 22 & 8 & 15 & 100 & 100 & 99 & 100\\
		\addlinespace[.2cm]
		\multicolumn{10}{l}{\quad\textit{Panel B: empirical rejection rates of mon. scheme \eqref{eq:page_cusum_test}}} \\ 
		acc. & 4 & 4 & 3 & 10 & 10 & 73 & 100 & 100 & 100\\
		conf. & 6 & 5 & 21 & 7 & 15 & 100 & 100 & 99 & 100\\
		\hline \hline
	\end{tabular} \medskip
	\end{center}
	\caption{Empirical rejection rates of CUSUM-based monitoring schemes based on the models' accuracies and confidences for the hypotheses \eqref{eq:null}, the models \LogReg, \NNI~ and \NNC~ and different data regimes.}
	\label{tab:real_data_cusum}
\end{table}

\end{document}